\documentclass[journal]{IEEEtran}
\usepackage{amsmath,amsfonts}
\usepackage{algorithmic}
\usepackage{array}
\usepackage[caption=false,font=normalsize,labelfont=sf,textfont=sf]{subfig}
\usepackage{textcomp}
\usepackage{stfloats}
\usepackage{url}
\usepackage{verbatim}
\usepackage{graphicx}
\usepackage{cite}
\usepackage[numbers]{natbib}
\usepackage{booktabs}
\usepackage{multirow}
\usepackage{makecell}
\usepackage{threeparttable}
\usepackage[table]{xcolor}
\usepackage[ruled,vlined,linesnumbered]{algorithm2e} 
\usepackage{hyperref}
\usepackage{marvosym}

\begin{document}

\title{
    Posterior Optimization with Clipped Objective for Bridging Efficiency and Stability in Generative Policy Learning
}

\author{
    % Anonymous Author
    Yuhui Chen, %~\IEEEmembership{Graduate Student Member,~IEEE,}
    Haoran Li\textsuperscript{\Letter}, %~\IEEEmembership{Member,~IEEE,}
    Zhennan Jiang, %~\IEEEmembership{Graduate Student Member,~IEEE,}
    Yuxing Qin, %~\IEEEmembership{Graduate Student Member,~IEEE,}
    Yuxuan Wan, %~\IEEEmembership{Graduate Student Member,~IEEE,}
    Weiheng Liu, %~\IEEEmembership{Graduate Student Member,~IEEE,}
    and Dongbin Zhao%,~\IEEEmembership{Fellow,~IEEE}
    
    \thanks{This work has been submitted to the IEEE for possible publication. Copyright may be transferred without notice, after which this version may no longer be accessible.}

    \thanks{This work was supported by the National Natural Science Foundation of China (NSFC) under Grants 62136008 and 62293545; in part by the Beijing Major Science and Technology Project under Contract no. Z251100008125023; and by the Suzhou Innovation and Entrepreneurship Leading Talents Programme – Innovation Leading Talent in Universities and Research Institutes under Grant ZXL2025310; and by Beijing Academy of Artificial Intelligence (BAAI).}
    
    \thanks{Yuhui Chen, Haoran Li, Zhennan Jiang, Yuxing Qin, and Dongbin Zhao are with the Institute of Automation, Chinese Academy of Sciences, Beijing 100190, China, and are also with the School of Artificial Intelligence, University of Chinese Academy of Sciences, Beijing 100049, China. Yuxuan Wan is with CFCS, School of Computer Science, Peking University, Beijing 100871. (Corresponding author: Haoran Li. E-mail: lihaoran2015@ia.ac.cn)}

}

\markboth{Journal of \LaTeX\ Class Files,~Vol.~1, No.~1, April~2026}%
{Shell \MakeLowercase{\textit{et al.}}: A Sample Article Using IEEEtran.cls for IEEE Journals}

% Remember, if you use this, you must call \IEEEpubidadjcol in the second

\maketitle

\begin{abstract}
Expressive generative models have advanced robotic manipulation by capturing complex, multi-modal action distributions over temporally extended trajectories. However, fine-tuning these policies via RL remains challenging due to instability and sample inefficiency. We introduce Posterior Optimization with Clipped Objective (POCO), a principled RL framework that formulates policy improvement as a posterior inference problem tailored for temporal action chunks. Through an Expectation-Maximization procedure, POCO distills a reward-weighted implicit posterior into the policy without likelihood estimation. Furthermore, POCO adopts an offline-to-online paradigm that anchors online exploration to pre-trained priors, and its model-agnostic design scales to fine-tune large VLA models without architectural modifications. Evaluations across 7 simulation benchmarks and 4 contact-rich real-world tasks demonstrate that POCO prevents catastrophic policy collapse, outperforms SOTA baselines, and achieves a 96.7\% success rate on real-world tasks. Videos are available at our project website \href{https://cccedric.github.io/poco/}{https://cccedric.github.io/poco/}.
\end{abstract}


\section{Introduction}
\IEEEPARstart{R}{obotic} manipulation is undergoing a critical paradigm shift. To tackle the multi-modal action distributions inherent in complex, contact-rich physical environments, policy representations have rapidly evolved from traditional uni-modal Gaussian distributions to highly expressive generative models \cite{urain2026survey, chen2022offline}, such as diffusion models \cite{dp}, flow matching \cite{fm}, and billion-parameter Vision-Language-Action (VLA) models \cite{pi05, gr00tn1}. Crucially, to prevent jitter and ensure smooth execution in high-frequency control, these generative architectures typically predict sequences of future movements rather than a single step, a paradigm known as temporal action chunking \cite{act}. At the same time, reinforcement learning (RL) has emerged as a powerful paradigm for equipping these models with complex decision-making capabilities, demonstrating remarkable success in high-dimensional, continuous, dexterous manipulation tasks \cite{rlinfvla, wu2025assembly, hu2023grasping}. However, applying RL for these generative models to continuously adapt and improve in the real world still faces significant bottlenecks \cite{hilserl}.

\begin{figure}[t]
    \centering
    \includegraphics[width=\columnwidth]{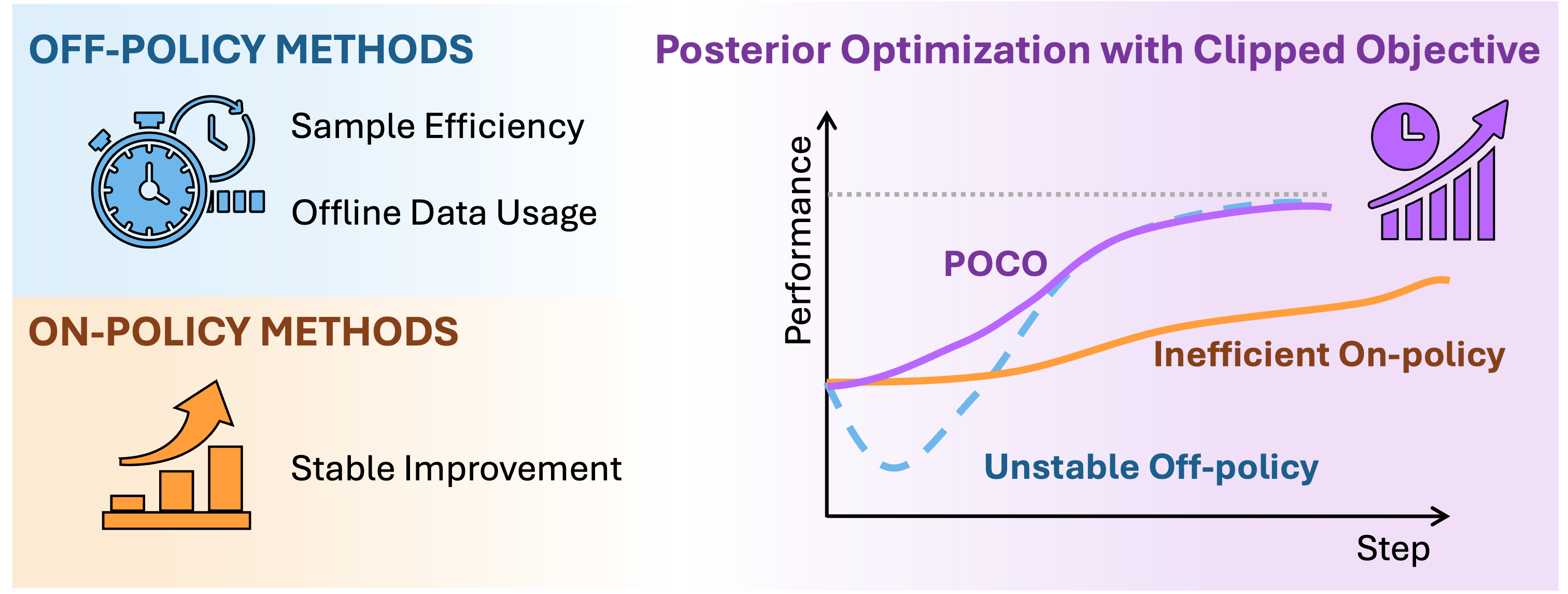}
    \caption{\textbf{Conceptual overview comparing typical RL paradigms against POCO.} (Left) Off-policy methods enable efficient data reuse, while on-policy methods ensure stability. (Right) Schematic performance curves show that POCO is designed for sample-efficient, stable improvement.}
    \label{fig:illustration}
\end{figure}

Because learning purely from online interaction is impractically slow \cite{sop}, real-world robotic systems usually rely on the offline-to-online paradigm \cite{lee2021off2on, awac, rl100}, where policies are first pre-trained on offline data and then fine-tuned via online rollouts \cite{parl, zhang2026traversability}. However, fine-tuning generative policies is typically trapped in a stability-efficiency dilemma, as illustrated in Fig. \ref{fig:illustration}. On the one hand, off-policy methods are highly sample-efficient because they can reuse offline data. Yet they backpropagate noisy Q-gradients directly into the policy network. This issue is exacerbated when scaling to high-dimensional action chunks, as the accumulating temporal errors across the predicted chunk inevitably lead these updates to suffer from severe out-of-distribution (OOD) value over-estimation \cite{cql, fql}. This destroys the pre-trained priors learned from human demonstrations, leading to catastrophic policy collapse and unpredictable, dangerous behavior on physical hardware. On the other hand, on-policy methods enforce trust regions \cite{ppo, deepseekr1} to ensure stable improvement without collapsing, but they are prohibitively sample-inefficient in real-world continuous control, as they have to learn solely from on-policy rollouts \cite{reinflow}.

% \IEEEpubidadjcol

To break this dilemma and prevent the destruction of pre-trained priors, we argue that policy improvement is more naturally understood as a posterior inference problem rather than direct parameter optimization, as seen in most existing RL methods \cite{ppo, sac}. From this perspective, the current policy serves as a prior over action chunks, Q-values provide evidence, and policy improvement distills a better action distribution by reweighting the prior via an Expectation-Maximization (E-M) procedure. Theoretically, this framework elegantly bridges the gap: by establishing a variational upper bound on sequence distributions, it leverages sample efficiency by using off-policy data while enforcing conservative prior regularization to anchor pre-trained behaviors. However, while existing inference-based RL methods \cite{mpo, vmpo} yield theoretically monotonic improvement, they critically rely on explicit policy likelihood, such as evaluating the exact single-step action likelihood to construct and project the posterior. This requirement poses a critical challenge: for most generative architectures, including VLA models, exact likelihood evaluation of high-dimensional, multi-step action chunks is mathematically intractable or computationally impractical for real-time robotic control.

In this paper, we bridge this gap by introducing Posterior Optimization with Clipped Objective (POCO). By discarding explicit likelihood estimation, POCO constructs an implicit posterior from a set of importance-sampled action chunks, each weighted by its corresponding chunk-level Q-value. It then safely distills these high-value behaviors back into the policy using a clipped surrogate objective. This update mechanism is stable, preventing the catastrophic physical collapse usually seen in gradient-based methods. Furthermore, it seamlessly supports likelihood-free Flow Matching policies and large-scale foundation models without altering their underlying architectures. The primary contributions of this work could be summarized as follows:
\begin{enumerate} 
    \item We propose POCO, a principled RL framework that theoretically formulates generative policy improvement as a likelihood-free posterior inference problem. By integrating an Implicit E-M procedure with a robust clipped regression mechanism, our framework enables stable, posterior-guided fine-tuning for temporal action chunks, eliminating the need for explicit likelihood estimation and preventing catastrophic policy collapse.

    \item We utilize an offline-to-online training paradigm that bridges pre-trained generative priors with online environmental exploration. Specifically, our framework safely anchors online exploration to the pre-trained prior via posterior inference. And this model-agnostic design is scalable, can be directly applied to fine-tune large-scale VLA models without architectural modifications.

    \item We systematically evaluate our method across 7 simulated and 4 real-world benchmarks. Empirical results demonstrate that POCO outperforms state-of-the-art (SOTA) baselines in both sample efficiency and performance improvement, achieving a 96.7\% real-world success rate within 50K online training steps. 
\end{enumerate}

\section{Related Work}
\subsection{Expressive Policies for Robotic Manipulation}
Recent advancements in robotic manipulation have increasingly relied on expressive generative models to capture the multi-modal, high-dimensional action distributions inherent in complex, contact-rich tasks. Traditional policy representations, such as Gaussian models, often fail to model the intricate behaviors required for dexterous manipulation \cite{dp}. Consequently, researchers have turned to generative architectures. Diffusion policies \cite{dp3, dp, wang2025hierarchical, huang2025improving}, consistency policies \cite{cpql, cp}, and flow-matching policies \cite{fm, fql, ding2025fast} have demonstrated remarkable success in behavior cloning (BC) from expert demonstrations, largely due to their ability to represent complex distributions. Furthermore, the integration of vision and language into these generative frameworks has led to the development of large-scale VLA models \cite{rt2, openvlaoft, xvla, gr00tn1, pi05}. While these models exhibit strong reasoning and semantic priors, adapting them to downstream tasks with only supervised fine-tuning (SFT) remains challenging \cite{conrft, simplevlarl}. Compared to existing approaches, our work introduces an efficient RL framework specifically tailored for generative policies. By formulating policy improvement as a likelihood-free posterior inference problem, we integrate online exploration with pre-trained priors, accelerating stable downstream adaptation.

\subsection{Real-World RL Implementations}
Deploying RL on physical robots is fundamentally hindered by the sample inefficiency and safety risks of pure online exploration \cite{serl, hilserl}. To mitigate this, the offline-to-online paradigm \cite{lee2021off2on, cql, iql, calql, awac, rl100} pre-trains policies on static datasets before online fine-tuning. However, offline-to-online fine-tuning faces a strict trade-off between stability and efficiency across two main approaches. First, off-policy methods \cite{hilserl, conrft, zhao2025real, zhang2026traversability} are sample-efficient but unstable. This instability stems from Q over-estimation on OOD states and the direct backpropagation of noisy Q-gradients, which induces catastrophic forgetting of the pre-trained priors \cite{calql}. Consequently, these approaches often require costly human-in-the-loop (HIL) interventions \cite{hgdagger, hilserl} to recover from failures and guide the learning process. Second, on-policy methods \cite{ppo, dppo, reinflow, rlinfvla} enforce strict trust regions to stabilize updates but suffer from prohibitive sample inefficiency in real-world continuous control. Advantage-conditioned methods \cite{dd, pi06} transform RL into a supervised fine-tuning problem to avoid noisy Q-gradients. Yet they still rely on continuous, costly HIL corrections during online rollouts to prevent policy degeneration. Alternative safe fine-tuning methods bypass direct weight updates by steering the generative process, either through altering inference noise \cite{dsrl} or applying value guidance \cite{vgps}. While these approaches provide stability, the maximum performance of the policy is limited because the underlying model weights remain unchanged. In contrast, our method bridges the gap between off-policy efficiency and on-policy stability. By formulating policy improvement as an E-M inference problem with a clipped surrogate objective, our method achieves stable, sample-efficient policy improvement through online rollouts without intervention actions.

\begin{figure*}[ht]
    \centering
    \includegraphics[width=\textwidth]{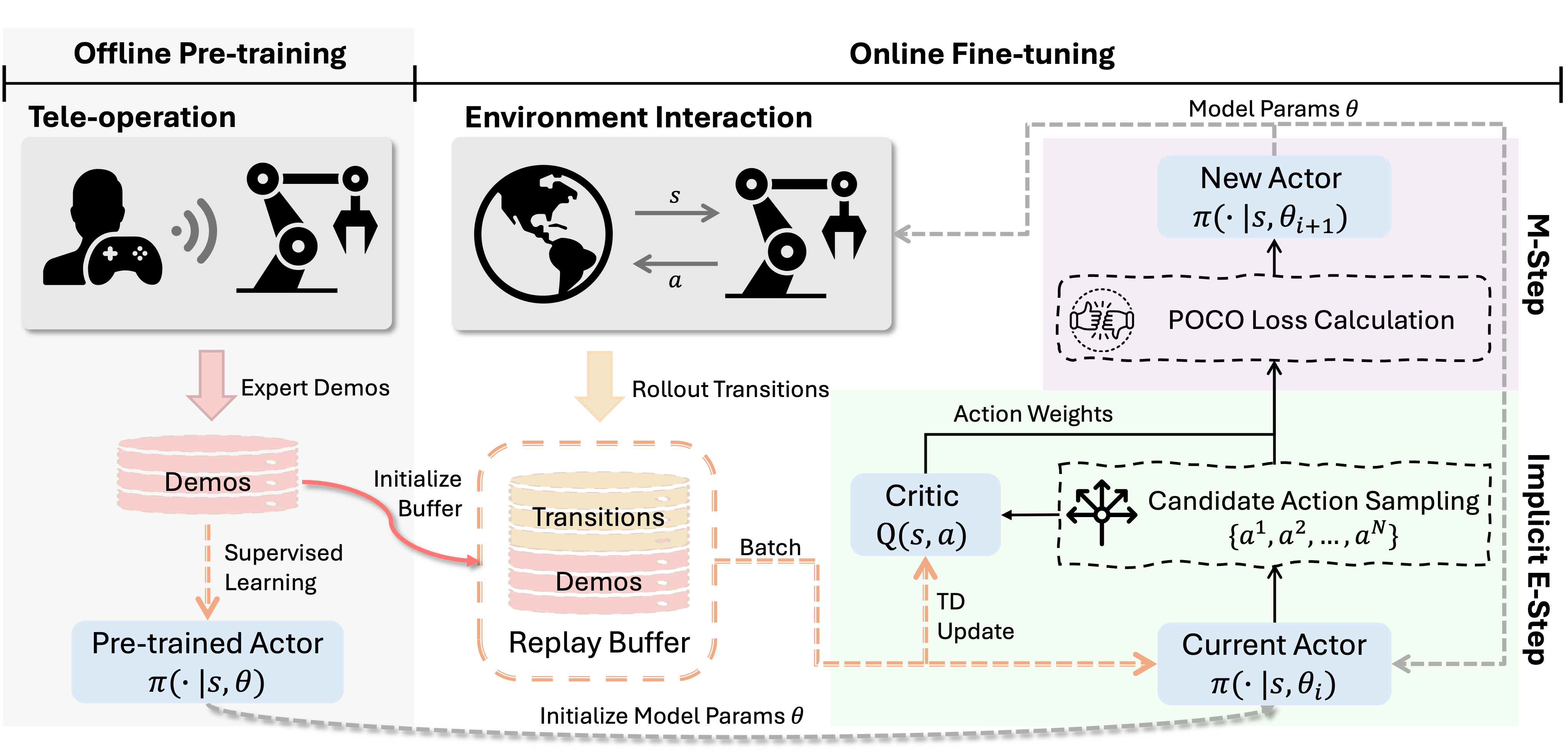}
    \caption{\textbf{Overview of our proposed framework.} The learning paradigm consists of two main stages. (Left) First, the policy is pre-trained through supervised learning using expert demonstrations collected via tele-operation. (Right) Second, during online fine-tuning, the policy interacts with the environment and improves through an iterative E-M procedure. In the Implicit E-step, multiple candidate actions are sampled from the current actor and evaluated by the critic to obtain importance weights. In the M-step, these weighted samples are utilized to compute the POCO loss, updating the parameters to obtain the new actor.}
    \label{fig:method}
\end{figure*}

\section{Background}
\subsection{Problem Setup}
We consider the problem of finding an optimal policy $\pi$ for a discounted RL problem, formally characterized by a Markov decision process (MDP). The MDP consists of $\mathcal{M} = (\mathcal{S}, \mathcal{A}, \mathcal{P}, r, \rho, \gamma)$, where $s \in \mathcal{S}$ denotes the state space and $a \in \mathcal{A}$ denotes the action space. $\mathcal{P}(s'|s, a)$ is the environmental transition probabilities that depend on system dynamics, and $\rho(s)$ denotes the initial state distribution. $r(s, a)$ and $\gamma \in (0, 1)$ are reward function and discount factor. 

A trajectory is defined as $\tau_{\pi}=\{(s_0,a_0), \dots, (s_H, a_H)\}$ sampled by the policy $\pi$ executing these generated action chunks sequentially, where $H$ represents the maximum episode step of a trajectory. Typically, the goal of optimizing the policy is maximizing the cumulative expected value of the reward, denoted as:
\begin{equation}
    \begin{aligned}
    V^{\pi}(s_t)=\mathbb{E}_{\pi}[\sum_{k=0}^{H-t}\gamma^kr(s_{t+k}, a_{t+k})|s_t]
    \end{aligned}
\end{equation}
The Q-function of a given policy $\pi$ evaluates the expected return of executing the action $a_t$ starting from state $s_t$. This is formally denoted as:
\begin{equation}
    \begin{aligned}
    Q^{\pi}(s_t,a_t)=\mathbb{E}_{s_{t+1}\sim\mathcal{P}}[r(s_t,a_t)+\gamma V^\pi(s_{t+1})]
    \end{aligned}
    \label{eq:standard_q}
\end{equation}

To accommodate high-frequency robotic control and temporal action chunking, we extend the standard single-step action formulation. Let an action chunk of horizon $T$ at time $t$ be defined as $\mathbf{a}_{t:t+T} = [a_t, a_{t+1}, \dots, a_{t+T-1}] \in \mathcal{A}^T$. The policy $\pi(\mathbf{a}_{t:t+T}|s_t, \theta)$ parameterized by $\theta$ is assumed to specify a probability distribution over sequences of future actions given the current state. 

\subsection{Policy Improvement as Inference}
To achieve the reward-maximizing goal defined in the MDP while maintaining the stability required for training expressive generative models, we can recast the policy optimization process as a probabilistic inference problem\cite{mpo}. RL ultimately seeks policies that make the expected outcomes more likely. To formalize this connection between the MDP and probabilistic inference, we define a binary optimality variable $\mathcal{O}$, where $\mathcal{O}=1$ denotes that a trajectory is expected, such as reaching a goal, accumulating high reward, or satisfying task constraints. A finite-horizon undiscounted RL problem can be cast as an inference problem by defining a likelihood model that connects the aforementioned MDP reward and optimality. Following the maximum-entropy framework, we specify:
\begin{equation}
    \begin{aligned}
    p(\mathcal{O}=1|\tau)\propto \exp(\sum_t(r_t/\eta))
    \end{aligned}
\end{equation}
where $\eta$ is a temperature parameter. This likelihood model captures the intuition that trajectories with larger cumulative reward should be exponentially more likely to be optimal. Under policy $\pi$, the probability that a sampled trajectory is optimal yields:
\begin{equation}
    \begin{aligned}
    \log p_{\pi}(\mathcal{O}=1) = \log\int p_{\pi}(\tau)p(\mathcal{O}=1|\tau)d\tau 
    \end{aligned}
\end{equation}

Note that directly maximizing this quantity is intractable due to the integral over all trajectories. We therefore introduce an auxiliary variational distribution $q(\tau)$. Applying importance sampling and Jensen’s inequality yields an evidence lower bound (ELBO) that serves as our basic optimization objective:
\begin{equation}
    \begin{aligned}
    \log p_{\pi}(\mathcal{O}=1)&=\log\int\frac{q(\tau)p_{\pi}(\tau)p(\mathcal{O}=1|\tau)}{q(\tau)}d\tau \\
    &\geq \int q(\tau)\Big(\log p(\mathcal{O}=1|\tau)+\log\frac{p_{\pi}(\tau)}{q(\tau)}\Big)d\tau \\
    &\equiv \frac{1}{\eta}\mathbb{E}_{q}\Big[\sum_t r_t\Big] - D_{\text{KL}}(q(\tau)||p_{\pi}(\tau)) \\
    &= \mathcal{J}(q, \pi)
    \end{aligned}
    \label{eq:elbo}
\end{equation}

\subsection{Conditional Flow Matching}
Standard policy optimization algorithms often rely on explicitly evaluating the action likelihood $\pi(a|s)$, which becomes computationally intractable for highly expressive generative models. To accommodate such likelihood-free architectures, we instantiate our framework using Flow Matching (FM) \cite{fm} as a representative generative policy built on continuous normalizing flows. Notably, the training objective illustrated below can be interpreted as a generalized supervised loss, making it naturally compatible with our proposed method.

Instead of directly outputting a deterministic action or relying on intractable likelihood estimation, a flow matching policy learns a time-dependent vector field that continuously transforms a simple prior distribution, typically a standard Gaussian $p_0(a)=\mathcal{N}(0, I)$, into the target data distribution $p_1(a|s)$ at $m=1$, where $m \in [0,1]$ represents the diffusion timesteps.

The continuous transformation of the action variable is governed by an Ordinary Differential Equation (ODE):
\begin{equation}
    \frac{d a^m}{dm} = v_\theta(a^m, m, s)
\end{equation}
where $v_\theta$ is a neural network parameterized by $\theta$ and $a^m$ represents the intermediate state of the action at diffusion timestep $m$. During the inference phase, an action is generated by sampling initial noise $a^0 \sim \mathcal{N}(0, I)$ and simulating the ODE forward to $m=1$ conditioned on the environment state $s$.

To train the vector field $v_\theta$ efficiently without the computational bottleneck of simulating the ODE through time during training, the policy constructs analytically tractable target probability paths $p_m(a^m|a^1)$ and their corresponding conditional vector fields $u_m(a^m|a^1)$. In our implementation, we use a linear probability path between the noise and the target action:
\begin{equation}
    a^m = (1-m)a^0 + m a^1
\end{equation}
where $a_1$ is the target action sampled from the dataset distribution, and $a_0 \sim \mathcal{N}(0, I)$ is the prior noise. Under this formulation, the target vector field simplifies to a constant velocity: $u_m(a^m|a^1) = a^1-a^0$.

Consequently, the policy network $v_\theta$ can be optimized using a supervised loss objective. This objective minimizes the expected Mean Squared Error (MSE) between the predicted vector field and the target vector field:
\begin{equation}
    \begin{aligned}
    \mathcal{L}_{\text{BC}}(\theta)=\mathbb{E}_{a^1 \sim \mathcal{D}}\Big[\|v_\theta(a^m,m,s)-(a^1-a^0)\|^2\Big]
    \end{aligned}
\end{equation}
where $m \sim \mathcal{U}[0,1]$ and $a^0 \sim \mathcal{N}(0,I)$. By minimizing $\mathcal{L}_{\text{BC}}$, the policy efficiently learns to approximate the marginal vector field. This formulation provides a stable, likelihood-free training objective that effectively captures the multi-modal action space required for dexterous robotic manipulation.

\section{Method}
In this section, we introduce Posterior Optimization with Clipped Objective (POCO), a framework for policy improvement, as illustrated in Fig. \ref{fig:method}. Unlike standard inference-based RL methods that operate on single-step actions, POCO formulates the posterior inference over action chunks. This perspective fits the output structure of common generative robotic policies, yielding a stable update rule that preserves temporal consistency in high-frequency continuous control.

For notational brevity in the subsequent sections, we use $\vec{\mathbf{a}}_t$ and $\vec{\mathbf{a}}'_t$ to denote the current action chunk $a_{t:t+T}$ and the subsequent action chunk $a_{t+T:t+2T}$, respectively. 

\subsection{Posterior Over Actions}
To make the optimization of ELBO in Eq. \ref{eq:elbo} tractable without relying on an explicit model of environmental dynamics, we must first decompose the trajectory-level distribution. Under the standard step-wise inference paradigm, the variational distribution is factorized over individual time steps: $q(\tau)=p(s_0)\prod_{t=0}^{H} p(s_{t+1}|s_t, a_t) q(a_t|s_t)$. Substituting this into the ELBO, the trajectory-level divergence decomposes into a sum of divergences:
\begin{equation}
    \begin{aligned}
    D_{\text{KL}}(q(\tau)||p_\pi(\tau)) = \sum_{t=0}^H\mathbb{E}_{q} [D_{\text{KL}}(q(a_t|s_t) \| \pi(a_t|s_t, \theta))]
    \end{aligned}
    \label{eq:kl}
\end{equation}

While this step-wise decomposition strictly aligns with the underlying MDP dynamics, making sequential decisions at a single-step granularity in high-frequency robotic control inherently leads to jittery execution and compounding errors. Leveraging the chain rule of relative entropy, the KL divergence between the joint distributions of the action chunks can be exactly decomposed into the divergence of the single-step marginal and the expected divergence of the subsequent conditional distributions:
\begin{equation}
    \begin{aligned}
    &D_{KL}(q(\vec{\mathbf{a}}_t|s_t)||\pi(\vec{\mathbf{a}}_t|s_t, \theta))=D_{KL}( q(a_t|s_t)||\pi(a_t|s_t, \theta) )\\
    &+\mathbb{E}_{a_t\sim q}[D_{KL}(q(a_{t+1:t+T}|a_t,s_t)||\pi(a_{t+1:t+T}|a_t, s_t,\theta))]\\
    \end{aligned}
\end{equation}

Since the KL divergence is non-negative, optimizing the chunk-level divergence could serve as a variational upper bound for minimizing the single-step divergence:
\begin{equation}
    \begin{aligned}
    &D_{KL}( q(a_t|s_t)||\pi(a_t|s_t,\theta))\le D_{KL}(q(\vec{\mathbf{a}}_t|s_t)||\pi(\vec{\mathbf{a}}_t|s_t, \theta))
    \end{aligned}
\end{equation}

By lifting the action space to temporal chunks and minimizing this upper bound, and incorporating the discount factor $\gamma$ into the likelihood model to ensure practical algorithmic convergence, we adopt a chunk-level Maximum A Posteriori (MAP) perspective. Treating the policy parameters $\theta$ as a random variable with a prior $p(\theta)$ naturally incorporates a regularization term, leading to the final unified objective:
\begin{equation}
    \begin{aligned}
    \mathcal{J}(q,\theta)=&\mathbb{E}_q\Big[\sum_{t=0}^{H} \gamma^t[r_t-\eta D_{\text{KL}}(q(\vec{\mathbf{a}}_t|s_t)||\pi(\vec{\mathbf{a}}_t|s_t,\theta))]\Big] \\
    &+\log p(\theta)
    \label{eq:elbo_final}
    \end{aligned}
\end{equation}

This formulation justifies an E-M procedure tailored for sequence generation: an E-step that constructs a non-parametric target distribution over action chunks weighted by values, followed by an M-step that updates $\theta$ under the trust-region constraint implied by $p(\theta)$.

\subsubsection{E-step (Chunk-Level Posterior Evaluation)}
At the $i$-th iteration of the E-M process, the E-step computes the optimal non-parametric posterior $q_i(\vec{\mathbf{a}}_t|s_t)$ that maximizes the state-wise ELBO in Eq. \ref{eq:elbo_final}. Here, the subscript $i$ denotes the current algorithm iteration. Solving this variational problem independently for each state yields a closed-form solution:
\begin{equation}
    q_i(\vec{\mathbf{a}}_t|s_t) \propto \pi(\vec{\mathbf{a}}_t|s_t,\theta_i)\exp\Big(\frac{Q_{\pi_i}(s_t,\vec{\mathbf{a}}_t)}{\eta}\Big)
    \label{eq:closed_form_q}
\end{equation}

Detailed mathematical proof is provided in Appendix \ref{apx:closed_form_q}. Note that this posterior acts as a reward-weighted refinement of the current policy, shifting probability mass towards the action distribution with higher expected returns. To evaluate this temporally extended evidence, the standard critic in Eq. \ref{eq:standard_q} is inadequate because it assumes the policy intervenes at every time step. In contrast, temporal action chunking executes $\vec{\mathbf{a}}_t$ in an open-loop manner for $T$ steps, requiring us to directly evaluate the expected return of the entire chunk. Crucially, this chunk-level formulation acts as a multi-step return estimator. Training with a $T$-step horizon accelerates back-propagation of sparse rewards, effectively alleviating the long-horizon credit assignment problem and facilitating faster critic convergence.

To achieve this, we introduce a chunk-level critic $\mathcal{Q}_\phi(s_t, \vec{\mathbf{a}}_t)$ with parameter $\phi$. Because the intermediate rewards over the subsequent $T$ steps depend solely on the environmental transition dynamics under $\vec{\mathbf{a}}_t$, we can extend the standard Bellman expectation equation to a multi-step formulation. We train this chunk-level critic by minimizing the $T$-step temporal difference (TD) residual on the replay buffer $\mathcal{D}$:
\begin{equation} 
    \begin{aligned} 
    \mathcal{L}_{\mathcal{Q}}(\phi)&=\mathbb{E}_{(s_t,\vec{\mathbf{a}}_t,r_{t:t+T},s_{t+T})\sim\mathcal{D}}\Big[\Big(\mathcal{Q}_\phi(s_t,\vec{\mathbf{a}}_t)\\
    -&\Big(\sum_{k=0}^{T-1}\gamma^k r_{t+k}+\gamma^T\mathbb{E}_{\vec{\mathbf{a}}'_t\sim \pi(\cdot|s_{t+T})}[\mathcal{Q}_{\bar{\phi}}(s_{t+T},\vec{\mathbf{a}}'_t)]\Big)\Big)^2\Big]\\ 
    \end{aligned} 
    \label{eq:chunked_q}
\end{equation}
where $\mathcal{Q}_{\bar{\phi}}$ serves as a target network. The learned Q-values are then used to construct the targets $q_i(\vec{\mathbf{a}}_t|s_t)$ for the subsequent policy update.

\subsubsection{M-step (Policy Improvement)}
Given the posterior $q_i(\vec{\mathbf{a}}_t|s_t)$, the M-step updates the parametric policy $\pi_\theta$ to match this target distribution. We instantiate the Bayesian prior $\log p(\theta)$ as a trust-region constraint, $D_{\text{KL}}(\pi(\cdot|s_t, \theta_i) \| \pi(\cdot|s_t, \theta)) \le \epsilon$, to ensure stability. Solving this constrained optimization via Lagrange multipliers yields the following weighted objective:
\begin{equation} 
    \begin{aligned} 
    \mathcal{J}_{\text{M-step}}(\theta)=&\mathbb{E}_{s_t\sim\mu(s_t)}[\mathbb{E}_{a\sim q_i(\vec{\mathbf{a}}_t|s_t)}[\log\pi(\vec{\mathbf{a}}_t|s_t,\theta)]\\
    &-\eta D_{\text{KL}}(\pi(\cdot|s_t,\theta_i)||\pi(\cdot|s_t,\theta))]
    \end{aligned} 
    \label{eq:m_step_objective} 
\end{equation}
where $\eta$ here acts as the regularization coefficient. The first term projects the policy onto the high-return regions identified in the E-step, while the second term anchors the update to the previous policy $\pi_{\theta_i}$. This mechanism prevents policy collapse and ensures stable improvement by keeping the policy within the valid region of the critic. 

\subsection{Posterior Optimization with Clipped Objective}
While the aforementioned E-M procedure provides a principled and stable framework for policy improvement, its standard implementation relies on the explicit construction of the variational posterior $q(\vec{\mathbf{a}}_t|s_t)$. This requires the policy to possess a tractable likelihood for the projection step. Such a requirement becomes a significant bottleneck when employing expressive but likelihood-free policy representations, where the exact log-likelihood $\log \pi(\vec{\mathbf{a}}_t|s_t, \theta)$ is either intractable or computationally expensive to evaluate. To overcome these limitations, POCO decouples policy improvement from explicit likelihood estimation by characterizing an implicit posterior and designing a sample-based improvement operator.

\subsubsection{Implicit E-step (Chunk-Level Posterior Evaluation via Weighted Samples)}
In the implicit E-step, instead of explicitly constructing the likelihood $q_i(\vec{\mathbf{a}}_t|s_t)$, we represent the posterior through $N$ candidate actions $\{\vec{\mathbf{a}}_t^j\}_{j=1}^N$ sampled from the current policy $\pi(\cdot|s_t, \theta_i)$. Following Eq. \ref{eq:closed_form_q}, each candidate action is assigned an unnormalized importance weight:
\begin{equation}
    \begin{aligned}
    w_j = \exp\Big(\frac{\mathcal{Q}(s_t, \vec{\mathbf{a}}_t^j)}{\eta}\Big)
    \end{aligned}
    \label{eq:weight}
\end{equation}

In practice, the exponential of large Q-values in Eq. \ref{eq:weight} might cause the loss to explode. We therefore normalize these weights, which yields a particle-based approximation of the reward-weighted posterior:
\begin{equation}
    \begin{aligned}
    \bar{w}_j=\frac{w_j}{\sum_{k=1}^N w_k}=\frac{\exp(\mathcal{Q}(s_t, \vec{\mathbf{a}}_t^j)/\eta)}{\sum_{k=1}^N \exp(\mathcal{Q}(s_t, \vec{\mathbf{a}}_t^k)/\eta)}
    \end{aligned}
    \label{eq:weight_normalized}
\end{equation}

This allows us to represent the posterior $q_i$ using only samples and their corresponding Q-values, bypassing the need for an explicit functional form.

\subsubsection{M-step (Policy Improvement with Clipped Objective)}
Given the weighted samples $\{\vec{\mathbf{a}}_t^j, \bar{w}_j\}$ from the Implicit E-step, we redefine the M-step objective in Eq. \ref{eq:m_step_objective} to support expressive policy representations. Specifically, we employ flow matching policies that converge to a regularized, optimal-transport-guided generative trajectory, characterized by smooth vector fields \cite{fm}. 

To support such likelihood-free policies, we must first map the likelihood-based objective into a tractable supervised learning space. As for generative modeling, the supervised training loss inherently minimizes a variational upper bound on the negative log-likelihood \cite{ddpm, score_matching}. Therefore, the supervised objective can be mathematically formulated as:
\begin{equation}
    \begin{aligned}
    \mathcal{L}_{\text{BC}, \theta}(\vec{\mathbf{a}}_t|s_t)\approx -\log\pi(\vec{\mathbf{a}}_t|s_t,\theta)+C
    \end{aligned}
    \label{eq:log_map}
\end{equation}
where $\mathcal{L}_{\text{BC},\theta}$ represents the supervised loss for parameter $\theta$, and $C$ represents an intractable normalization constant that is independent of the policy parameters $\theta$. Consequently, computing the difference between the supervised objectives of the current policy $\theta$ and the reference policy $\theta_i$ yields an approximation for the log-probability ratio:
\begin{equation}
    \begin{aligned}
    \log\frac{\pi(\vec{\mathbf{a}}_t|s_t,\theta_i)}{\pi(\vec{\mathbf{a}}_t|s_t,\theta)}\approx\mathcal{L}_{\text{BC},\theta}(\vec{\mathbf{a}}_t|s_t)-\mathcal{L}_{\text{BC},\theta_i}(\vec{\mathbf{a}}_t|s_t)
    \end{aligned}
\end{equation}

Substituting this approximation into the KL divergence term yields:
\begin{equation}
    \begin{aligned}
    D_{\text{KL}}(\pi(\cdot|s_t,\theta_i)||\pi(\cdot|s_t,\theta))=\mathbb{E}_{\vec{\mathbf{a}}_t \sim \pi(\cdot|s_t,\theta_i)}\Big[\log\frac{\pi(\vec{\mathbf{a}}_t|s_t,\theta_i)}{\pi(\vec{\mathbf{a}}_t|s_t,\theta)} &\Big] \\
    \approx\mathbb{E}_{\vec{\mathbf{a}}_t \sim \pi(\cdot|s_t,\theta_i)}[\mathcal{L}_{\text{BC},\theta}(\vec{\mathbf{a}}_t|s_t) - \mathcal{L}_{\text{BC},\theta_i}(\vec{\mathbf{a}}_t|s_t)&]
    \end{aligned}
\end{equation} 

In practical policy iteration, the reference policy parameters $\theta_i$ are treated as fixed constants during the optimization of the current policy $\theta$. Consequently, the expectation term $\mathbb{E}_{\vec{\mathbf{a}}_t \sim \pi(\cdot|s_t,\theta_i)}[\mathcal{L}_{\text{BC},\theta_i}\vec{\mathbf{a}}_t|s_t)]$ remains constant with respect to $\theta$ and does not contribute to the gradient of the objective function. By omitting this constant term, minimizing the KL divergence constraint mathematically reduces to minimizing the expected supervised loss of the current policy:
\begin{equation}
    \begin{aligned}
    &D_{\text{KL}}(\pi(\cdot|s_t,\theta_i)||\pi(\cdot|s_t,\theta)) \approx\mathbb{E}_{\vec{\mathbf{a}}_t \sim \pi(\cdot|s_t,\theta_i)}[\mathcal{L}_{\text{BC},\theta}(\vec{\mathbf{a}}_t|s_t)]
    \end{aligned}
\end{equation} 

In our training setting, we approximate the reference distribution $\pi(\cdot|s_t, \theta_i)$ using the data distribution from the replay buffer $\mathcal{D}$. Mathematically, this replaces the strict on-policy trust region with a BC prior regularization. This adaptation introduces a conservative bias that anchors the policy to the offline demonstrations, thereby actively preventing catastrophic forgetting of the pre-trained prior. Also, rather than optimizing the exact Lagrangian multiplier $\eta$, which is notoriously difficult to tune dynamically in deep RL, we decouple the objective and introduce a tunable hyperparameter $\beta$, termed the posterior guidance scale, to explicitly balance the surrogate term and the underlying trust-region regularization. Following the variational mapping, the M-step objective can be tractably optimized using the weighted supervised loss:
\begin{equation}
    \begin{aligned}
    \mathcal{J}_{\text{M-step}}(\theta)\approx&\mathbb{E}_{(s_t,\vec{\mathbf{a}}_t)\sim\mathcal{D}, \{\vec{\mathbf{a}}_t^j\}_{j=1}^N \sim \pi(\cdot|s_t,\theta_i)}\Big[\mathcal{L}_{\text{BC},\theta}(\vec{\mathbf{a}}_t|s_t)\\
    &+\beta\sum_{j=1}^N\bar{w}_j\mathcal{L}_{\text{BC},\theta}(\vec{\mathbf{a}}_t^j,s_t)\Big]
    \end{aligned}
\end{equation}

However, directly minimizing this objective may expose the policy to catastrophic manifold collapse. Highly-weighted OOD actions generated during the E-step typically lie far outside the expert data manifold. An MSE-based supervised objective would impose a large penalty, forcing the generative vector field $v_\theta$ to abruptly fit these distant outlier points. This violent update destroys the smooth topological structure of the pre-trained vector field.

\begin{algorithm}[t]
    \caption{Posterior Optimization with Clipped Objective}
    \label{alg:ippo}
    \DontPrintSemicolon
    \SetAlgoLined
    
    \SetKwInput{KwInput}{Input}
    \SetKwInput{KwParam}{Hyperparams}
    \SetKwComment{tcp}{}{}
    
    \KwInput{Actor $\pi_\theta$, Chunk-level Critic $\mathcal{Q}_\phi$, Replay Buffer $\mathcal{D}$}
    \KwParam{Chunk horizon $T$, Number of candidate actions $N$, Clipping threshold $\zeta$, Weight temperature $\eta$, Posterior guidance scale $\beta$}
    
    Initialize parameters $\theta, \phi$ randomly\;
    Initialized $\mathcal{D}$ with pre-collected demo trajectories\;
    \textcolor{gray!90}{\# Stage I: Offline Pre-training}\;
    \For{each offline training step}{
        Sample $\{s_{t}, \vec{\mathbf{a}}_t, r_{t:t+T}, s_{t+T}\}$ from $\mathcal{D}$\;
        Update the actor parameters $\theta$ by Eq. \ref{eq:offline_bc}\;
    }
    
    \textcolor{gray!90}{\# Stage II: Online Fine-tuning}\;
    \textbf{\# Start Interaction Thread:}\;
    \For{each interaction step $k$}{
        Observe current state $s_k$\;
        Sample and execute action $a_{k:k+T} \sim \pi(\cdot|s_k, \theta)$\;
        Store transitions $\{(s_t,a_t,r_t,s_{t+1})\}^T_{t=k}$ into $\mathcal{D}$
    }
    \textbf{\# Start Policy Learning Thread:}\;
    Set current actor $\theta_0\leftarrow\theta$ from Stage I\;
    \For{each online training step $i$}{
        Sample $\{s_t, \vec{\mathbf{a}}_t, r_{t:t+T}, s_{t+T}\}$ from $\mathcal{D}$\;
        \eIf{$i \le \text{critic warmup step}$}{
            \textcolor{gray!90}{\# Critic Warmup}\;
            Keep actor $\theta_i$ frozen\;
            Update the critic $\phi$ by Eq. \ref{eq:chunked_q}\;
        }{
            \textcolor{gray!90}{\# Implicit E-Step}\;
            Update the critic $\phi$ by Eq. \ref{eq:chunked_q}\;
            Sample candidate actions from current actor $\{\vec{\mathbf{a}}_t^j\}_{j=1}^N \sim \pi(\cdot|s_t, \theta_i)$\;
            Calculate weights for each action by Eq. \ref{eq:weight_normalized}\;
            \textcolor{gray!90}{\# M-Step}\;
            Update actor from $\theta_{i}$ to $\theta_{i+1}$ by Eq. \ref{eq:ippo}\;
            Copy parameter $\theta\leftarrow\theta_{i+1}$ for interaction\;
        }
    }
\end{algorithm}

To mitigate this, we introduce a clipped surrogate objective with a threshold $\zeta$. By clipping the weighted supervised objective at $\zeta$, we mathematically impose a maximum allowable geometric deformation per update step. It guarantees that the policy distills knowledge only from high-value actions within a structurally safe, $\zeta$-bounded neighborhood of the current vector field. The final objective for POCO is formulated as:
\begin{equation}
    \begin{aligned}
    \mathcal{J}_{\text{POCO}}(\theta)=&\mathbb{E}_{(s_t,\vec{\mathbf{a}}_t)\sim\mathcal{D}, \{\vec{\mathbf{a}}_t^j\}_{j=1}^N \sim \pi(\cdot|s_t,\theta_i)}\Big[\mathcal{L}_{\text{BC},\theta}(\vec{\mathbf{a}}_t|s_t)\\&+\beta\sum_{j=1}^N\bar{w}_j\text{clip}\Big(\mathcal{L}_{\text{BC},\theta}(\vec{\mathbf{a}}_t^j,s_t), 0, \zeta\Big)\Big]
    \end{aligned}
    \label{eq:ippo}
\end{equation}

This formulation illustrates that POCO is equivalent to performing posterior-guided policy improvement: the E-step identifies high-value trajectory segments through re-weighting, while the M-step distills this filtered knowledge back into the parametric policy $\pi_\theta$ using a likelihood-free objective. This allows our framework to keep policy stable while supporting the high expressivity required for complex robotic manipulation tasks.

\subsection{Offline-to-Online Training Paradigm}
To ensure safe and efficient learning in real-world environments, we adopt a two-stage training paradigm consisting of offline pre-training followed by online fine-tuning for POCO, as shown in Algorithm \ref{alg:ippo}.

\subsubsection{Offline Pre-training}
To equip the agent with high-quality initial behaviors before interaction in real-world environments, we perform supervised pre-training on the static dataset $\mathcal{D}$, initialized with pre-collected expert demonstrations. Specifically, we employ a DAgger-like data collection \cite{xvla} where human operators may intentionally demonstrate recovery actions from suboptimal states. Incorporating these corrective trajectories enables the policy to learn to recover during policy execution, thereby achieving robust initial prior in the offline stage. We optimize the flow-matching policy via supervised learning by minimizing:
\begin{equation}
    \label{eq:offline_bc} 
    \begin{aligned}
    \mathcal{J}^{\text{off}}(\theta)=\mathbb{E}_{(s_t,\vec{\mathbf{a}}_t)\sim\mathcal{D}}[\mathcal{L}_{\text{BC},\theta}(\vec{\mathbf{a}}_t|s_t)]
    \end{aligned}
\end{equation}

This supervised stage provides a strong policy initialization, allowing the agent to effectively explore high-reward regions during the online phase.

\subsubsection{Online Fine-tuning}
During the online phase, we continuously augment the dataset $\mathcal{D}$ with new transitions, i.e., $\mathcal{D} \leftarrow \mathcal{D} \cup \mathcal{D}_{\text{new}}$. To bridge the gap between offline priors and online dynamics, we adopt a two-stage update strategy. Initially, we address potential Q-value over-estimation and distribution-shift issues \cite{cql} by conducting a critic warm-up phase. In this stage, we employ a SARSA-style update \cite{richard1998rl}. Specifically, we freeze the trained actor parameters and train the chunk-level critic $\mathcal{Q}_\phi$ by estimating values of actions sampled from the fixed policy. This procedure ensures the critic learns conservative and stable value estimates, establishing a reliable guidance for subsequent policy improvements. Once the critic is warmed up, we unfreeze the actor and proceed to the standard iterative POCO updates. 

\section{Experiments}
In this section, we conduct a comprehensive empirical evaluation of our proposed framework across a diverse set of manipulation tasks. Our experiments are designed to answer five key questions regarding the performance of our method:
\begin{enumerate}
    \item \textbf{Sample Efficiency (Section\ref{sec:online_sim}):} Can our method learn complex policies from scratch with high sample efficiency under online settings?
    \item \textbf{Stability in Offline-to-Online Training (Section\ref{sec:off2on_sim}):} Does our method effectively mitigate catastrophic forgetting and enable stable fine-tuning in offline-to-online scenarios?
    \item \textbf{Real-World Applicability (Section\ref{sec:real}):} Can our method perform policy improvement robustly in real-world environments?
    \item \textbf{Ablation Study (Section\ref{sec:ablation}):} What is the impact of key algorithmic designs, such as the clipped surrogate objective and the implicit posterior weighting, on the effectiveness of our framework?
    \item \textbf{Scalability to VLA models (Section\ref{sec:vla}):} Is our method compatible with large-scale pre-trained VLA models to enhance their performance in real-world environments?
\end{enumerate}
To answer these questions, we compare our approach against SOTA baselines in both simulation benchmarks and real-robot environments.

\begin{figure}[t]
    \centering
    \includegraphics[width=\columnwidth]{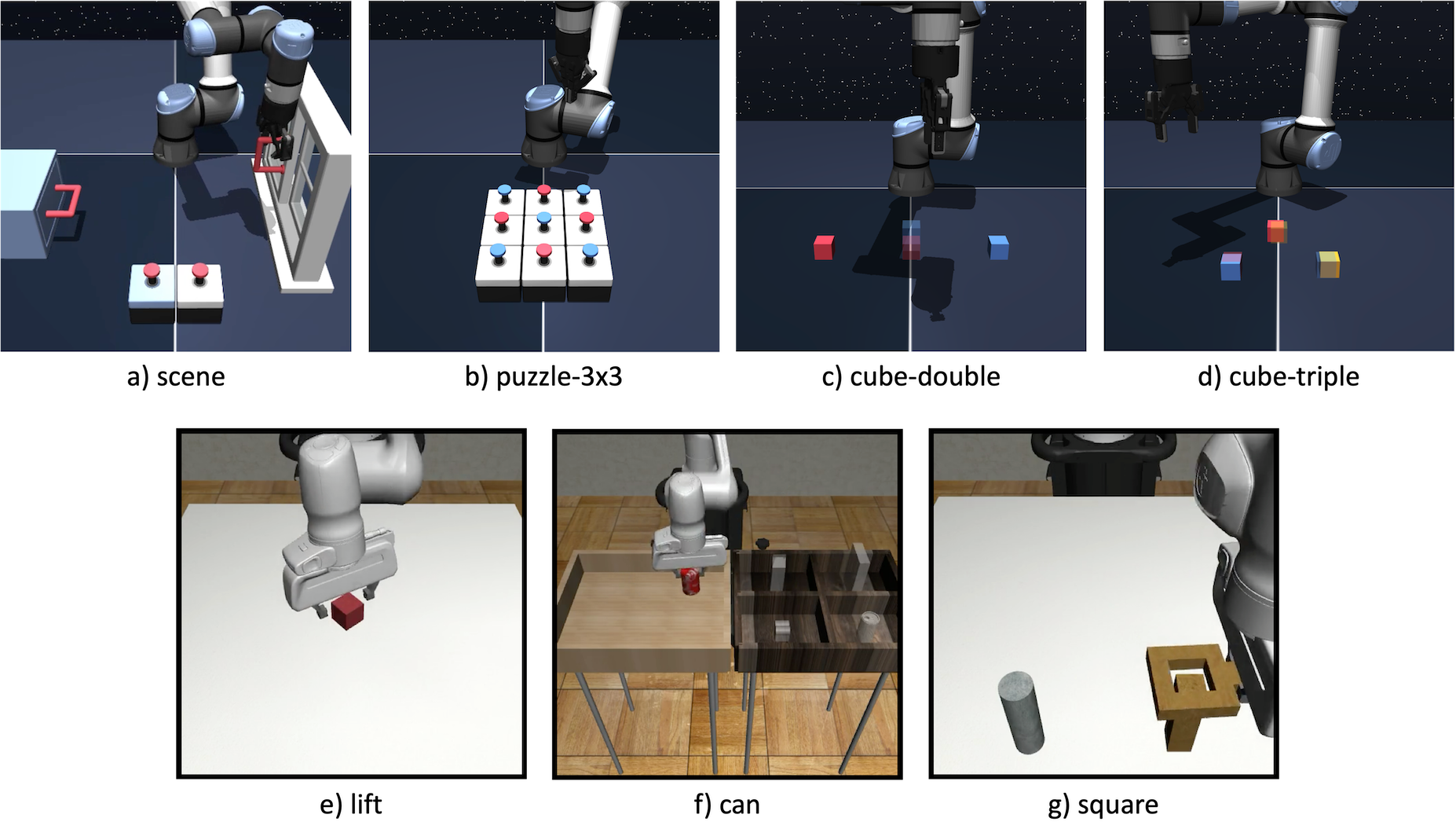}
    \caption{\textbf{Visualization of all simulation tasks.} The simulation tasks includes: a) \texttt{scene}, b) \texttt{puzzle-3x3}, c) \texttt{cube-double}, d) \texttt{cube-triple} from OGBench \cite{ogbench} and e) \texttt{lift}, f) \texttt{can}, g) \texttt{square} from RoboMimic \cite{robomimic}.}
    \label{fig:sim_via}
\end{figure}

\begin{table}[t]
\caption{\textbf{Metadata for offline datasets in simulation and real-world tasks.}}
\label{tab:data}
\centering
\resizebox{\columnwidth}{!}{
\begin{tabular}{cc|ccc}
    \toprule
    \multicolumn{2}{c}{\textbf{Tasks}} &\textbf{Num. of Trajs.} &\textbf{Num. of Trans.} &\textbf{Action Dimension} \\
    \midrule
    \multirow{7}{*}{\rotatebox{90}{\textbf{Simulation}}} &\texttt{scene} &1,000 &1,000,000 &5 \\
     &\texttt{puzzle-3x3} &1,000 &1,000,000 &5 \\
     &\texttt{cube-double} &1,000 &1,000,000 &5 \\
     &\texttt{cube-triple} &3,000 &3,000,000 &5 \\
    \cmidrule{2-5}
     &\texttt{lift} &300 &31,127 &7 \\
     &\texttt{can} &300 &62,756 &7 \\
     &\texttt{square} &300 &80,731 &7 \\
    \midrule
    \multirow{6}{*}{\rotatebox{90}{\textbf{Real-World}}} &\texttt{Pick Cube} &40 &3,471 &7 \\
     &\texttt{Route Cable} &30 &2,934 &7 \\
     &\texttt{Insert USB} &50 &3,880 &7 \\
     &\texttt{Assemble SSD} &40 &2,884 &7 \\
     &\texttt{Pick Pen} &50 &6,682 &7 \\
     &\texttt{Hang Keychain} &50 &5,996 &7 \\
    \bottomrule
\end{tabular}
}
\end{table}

\subsection{Implementation Details}
\subsubsection{Simulation Environments}
For the simulation benchmarks, we evaluate our algorithm on 7 sparse-reward robotic manipulation tasks of varying complexity. From OGBench\cite{ogbench}, we utilize "Task 1" of 4 domains: \texttt{scene}, \texttt{puzzle-3x3}, \texttt{cube-double}, and \texttt{cube-triple}. From RoboMimic\cite{robomimic}, we select 3 tasks: \texttt{lift}, \texttt{can} and \texttt{square}. Details for all simulation tasks are listed below.

\paragraph{OGBench}
For the \texttt{scene} and \texttt{puzzle-3x3} domains, we employ a sparse binary reward $r\in\{-1,0\}$, where a reward of $0$ is granted only upon task completion, and $-1$ otherwise. Detailed descriptions of each domain are provided below:
\begin{itemize}
    \item \texttt{scene}: A long-horizon task requires the agent to first unlock and open the drawer, then place the cube inside.
    \item \texttt{puzzle-3x3}: A $3\times 3$ grid-based task where pressing a button toggles its state (red/blue) and that of its cardinal neighbors. The task starts from all-blue buttons, requiring the agent to isolate and flip only the top-left button to red through a series of strategic moves.
    \item \texttt{cube-double/triple}: Multi-object manipulation domains with 2 or 3 cubes. In \texttt{cube-double} and \texttt{cube-triple}, the agent must rearrange all cubes to target poses. The reward is defined as $-n$, where $n$ denotes the count of cubes not yet at their goal positions. 
\end{itemize}

\paragraph{RoboMimic}
We utilize the Multi-Human (MH) datasets collected by multiple human operators to ensure diverse demonstrations. All tasks use binary task completion rewards where the agent receives a $1$ reward when the task is completed and a $-0.01$ reward otherwise. The tasks are defined as follows:
\begin{itemize}
    \item \texttt{lift}: A reaching and grasping task where the robot must lift a small cube off the tabletop.
    \item \texttt{can}: The agent is required to pick up a soda can and place it precisely into a designated container bin.
    \item \texttt{square}: A high-precision insertion task involving picking up a square nut and placing it onto a vertical rod, requiring high-fidelity control and spatial precision to prevent jamming during the contact phase.
\end{itemize}

All simulation tasks are visualized in Fig. \ref{fig:sim_via} and the metadata of the corresponding dataset is listed in Tab. \ref{tab:data}. For OGBench tasks, the 5D action space comprises End-Effector (EEF) Cartesian positions $(x, y, z)$, gripper yaw, and gripper opening. For RoboMimic tasks, the 7D action space controls the EEF pose $(x, y, z, roll, pitch, yaw)$ with the gripper state.

For a fair comparison, all algorithms used the flow matching policy\cite{fm} for simulation tasks, with a flow step of $10$. Both the actor and critic networks use a four-layer Multi-Layer Perceptron (MLP) with a hidden dimension of 512 per layer, employing Sigmoid Linear Unit (SiLU) as the activation function for the hidden layers, while the final output layer of both networks remains linear without any activation. Furthermore, to stabilize value estimation, the critic network incorporates Layer Normalization (LN) after the SiLU activation in each hidden layer.

\begin{figure}[t]
    \centering
    \includegraphics[width=0.9\columnwidth]{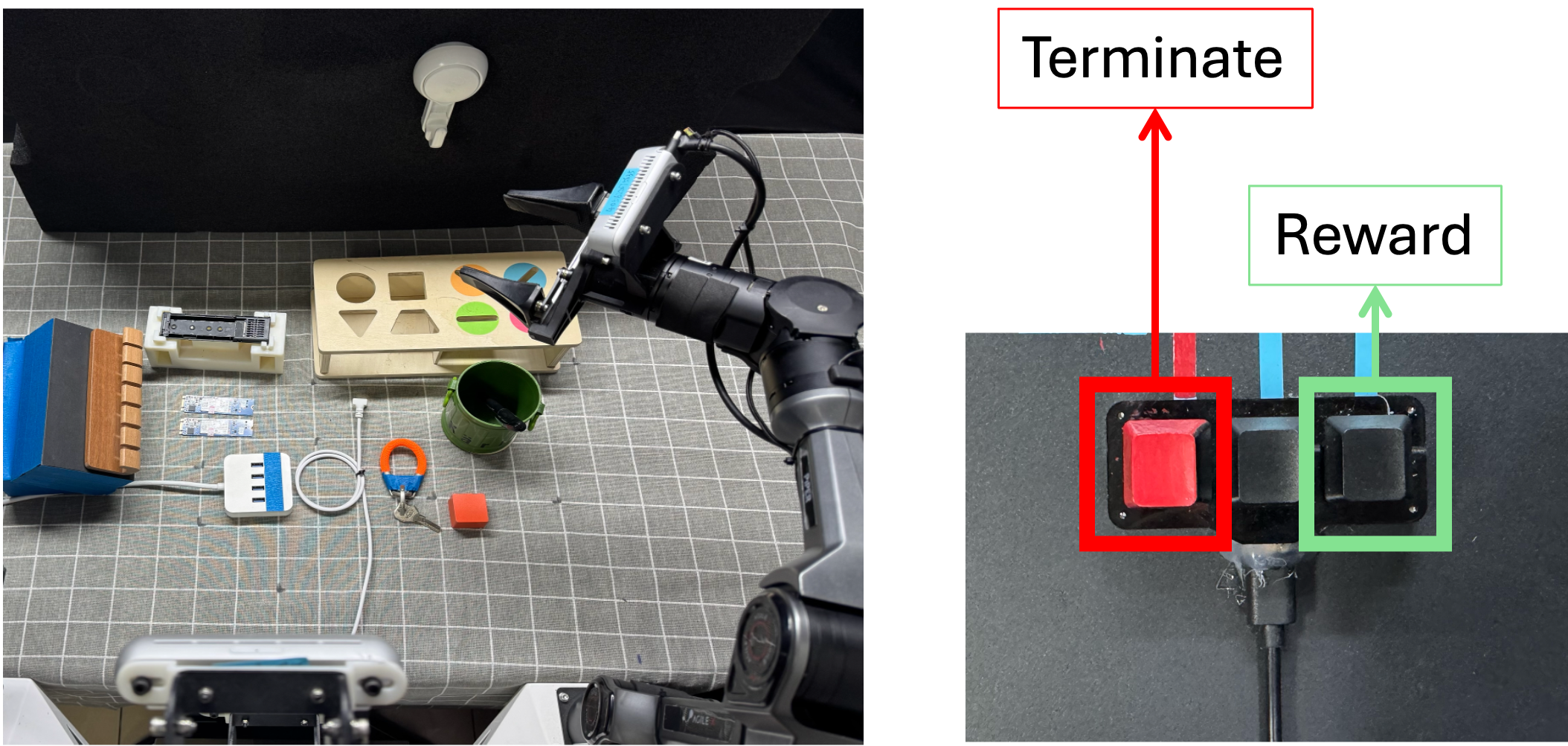}
    \caption{\textbf{Visualization of the real-world environment platform.} During execution, an operator presses the green button to assign a sparse reward upon success, or the red button to immediately terminate dangerous trajectories for hardware safety.}
    \label{fig:real_platform}
\end{figure}

\begin{figure}[t]
    \centering
    \includegraphics[width=0.9\columnwidth]{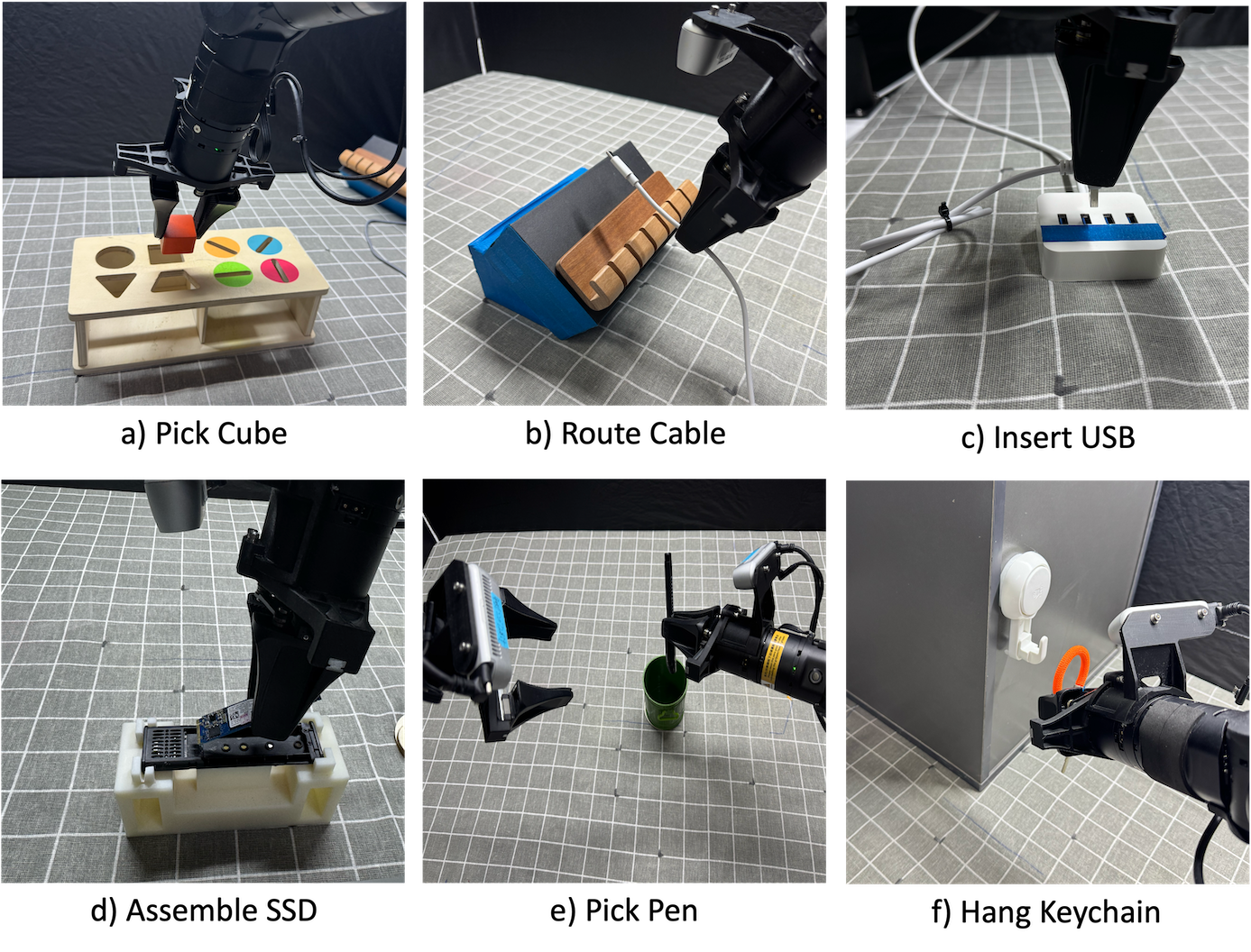}
    \caption{\textbf{Visualization of all real-world tasks.} The real-world tasks includes: a) \texttt{Pick Cube}, b) \texttt{Route Cable}, c) \texttt{Insert USB}, d) \texttt{Assemble SSD}, e) \texttt{Pick Pen}. The camera perspectives used in the experiments differ from those used for visualization.}
    \label{fig:real_vis}
\end{figure}

\subsubsection{Real-world Environments}
All real-world experiments are conducted using the AgileX Cobot Magic robotic system. The hardware setup features a 6 Degree-of-Freedom (DoF) robotic arm equipped with a 1 DoF parallel jaw gripper, providing a highly capable platform for dexterous manipulation. To collect the initial expert datasets, the system is integrated with a tele-operation interface for a collaborative robotic arm, enabling human operators to guide the robot to complete target tasks and record expert demonstrations. At the execution level, the low-level hardware controllers ensure smooth and responsive tracking of the policy's outputs, running at 15Hz. Furthermore, to accommodate the heavy computational overhead of the generative models while maintaining low-latency real-time control, we deploy a distributed system architecture. All compute-intensive policy training and inference processes are offloaded to a high-performance cloud server equipped with 8 NVIDIA H20 GPUs. Meanwhile, RGB and sensory data streaming, trajectory collection, and physical action execution are managed by a local computer that directly interfaces with the robot hardware over a local network.

The observation includes two RGB camera views (e.g., a front view and a wrist view) alongside the robot's proprioceptive state, which includes the current joint angles. The image observations are processed through a pre-trained frozen ResNet-10 backbone to extract low-dimensional spatial feature embeddings. These visual features are subsequently flattened and concatenated with the proprioceptive data to formulate the unified state representation. Then we employ a flow matching policy \cite{fm} with $10$ flow steps during inference. Both the actor and the Q network are parameterized as four-layer MLPs, with a hidden dimension of 512 per layer and SiLU as the activation function. The final output layer is linear, with no activation. Furthermore, to stabilize value estimation, the critic network incorporates LN after each activation.

For the offline pre-training stage, we initialize the offline datasets with 30 to 50 pre-collected successful trajectories per task. To capture natural and dexterous manipulation behaviors, these demonstrations are provided directly by human experts via tele-operation using a collaborative robotic arm. During the online fine-tuning phase, we employ a sparse binary reward formulation. Due to the inherent difficulty of automated state evaluation in unstructured physical environments, the reward signal is manually annotated by a human expert through a physical keypad (Fig. \ref{fig:real_platform}, right): the agent receives a reward of $1$ upon successful task completion and $0$ otherwise at the end of each episode. The tasks are defined as follows:

\begin{figure*}[ht]
    \centering
    \includegraphics[width=\textwidth]{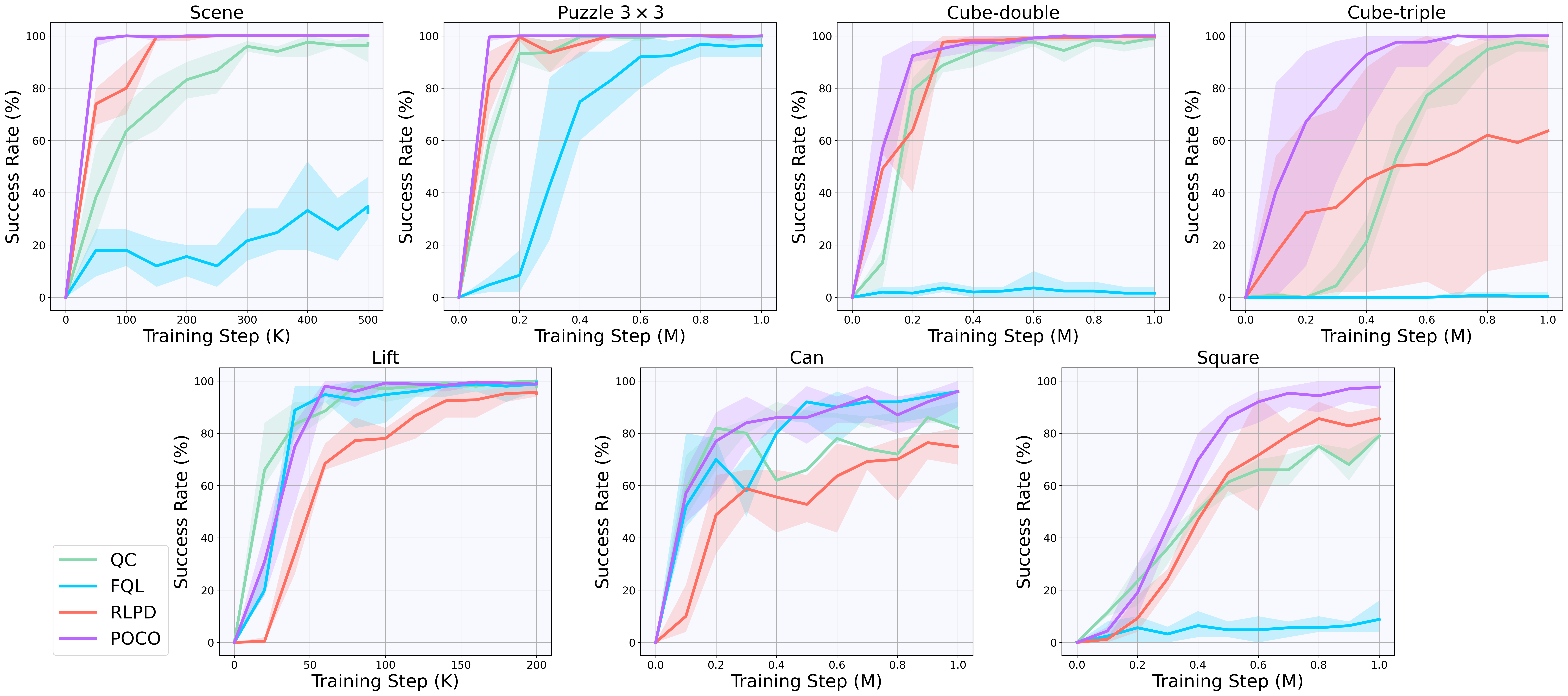}
    \caption{\textbf{Learning curves during online training for 7 simulation tasks.} Results are averaged over 50 trials per task at each evaluation step with 5 random seeds.}
    \label{fig:sim_online}
\end{figure*}

\begin{table*}[htbp]
    \caption{\textbf{Policy training details for online simulation experiments.} }
    \label{tab:sim_online_task_detail}
    \centering
    \resizebox{0.9\textwidth}{!}{
    \begin{tabular}{c|ccccccc}
        \toprule
        Parameter &\texttt{Scene} &\texttt{Puzzle 3x3} &\texttt{Cube-double} &\texttt{Cube-triple} &\texttt{Lift} &\texttt{Can} &\texttt{Square}\\
        \hline 
        Batch size &256 &256 &256 &256 &256 &256 &256\\
        Learning rate &3e-4 &3e-4 &3e-4 &3e-4 &3e-4 &3e-4 &3e-4\\
        $\gamma$ &0.99 &0.99 &0.99 &0.99 &0.99 &0.99 &0.99\\
        $T$ &5 &5 &5 &5 &5 &5 &5\\
        Critic warmup step &5,000 &5,000 &5,000 &5,000 &5,000 &5,000 &5,000\\
        $N$ &32 &32 &32 &32 &32 &32 &32\\
        $\eta$ &0.1 &0.1 &0.1 &0.1 &0.001 &0.001 &0.001\\
        $(\beta, \zeta)$ &$(1.0, 0.3)$ &$(1.0, 0.3)$ &$(1.0, 0.3)$ &$(1.0, 0.3)$ &$(1.0, 0.15)$ &$(1.0, 0.15)$ &$(1.0, 0.15)$\\
        \bottomrule
    \end{tabular}
    }
\end{table*}

\begin{itemize}
    \item \texttt{Pick Cube}: A peg-in-hole task where the agent need grasp a cube and insert it into a tightly fitted square hole. 
    \item \texttt{Route Cable}: A deformable object manipulation task requiring the agent to grasp a flexible cable and press it into a cable management.
    \item \texttt{Insert USB}: A high-precision task requiring the agent to pick up a USB connector and insert it into a port.
    \item \texttt{Assemble SSD}: A contact-rich assembly task where the agent must pick up an M.2 Solid-State Drive (SSD), align it at a specific oblique angle, and insert it into a motherboard slot. 
    \item \texttt{Pick Pen}: A pick-and-place task where the agent must grasp a thin, elongated pen from the desktop and insert it vertically into a cylindrical pen holder. The task description for the VLA model is ``Place the black pen in the green pen cup." 
    \item \texttt{Hang Keychain}: A precise spatial coordination task requiring the agent to grasp a keychain and thread its ring onto a fixed hook. The task description for the VLA model is ``Pick up the orange keychain and hang it onto the white hook." 
\end{itemize}

All real-world tasks are visualized in Fig. \ref{fig:real_vis} and the metadata of the corresponding dataset is listed in Tab. \ref{tab:data}. The 7D action space is parameterized by six delta joint position commands $(\Delta q_1, \Delta q_2, \dots, \Delta q_6)$, with a one-dimensional absolute state to control the gripper opening. 

\subsection{Sample Efficiency of Online Learning in Simulation}
\label{sec:online_sim}
To evaluate the sample efficiency of our approach, we compare 3 competitive methods with ours in the online setting across 7 simulation tasks. The specific details of the policy training for each task are in Tab. \ref{tab:sim_online_task_detail}. All results are provided with the same set of expert demonstrations to facilitate learning:
\begin{itemize}
    \item RLPD\cite{rlpd} is an off-policy method that maintains a separate buffer for offline demonstrations and mixes them with data from the replay buffer under a fixed ratio (e.g., 50\%) during updates to accelerate online exploration.
    \item QC\cite{qc} that utilized chunked Q function and Max-K action selection for robust and efficient policy update.
    \item FQL\cite{fql} is also an off-policy method that achieves high sample efficiency by directly back-propagating Q function gradients into the flow matching policy network.
\end{itemize}
We report the average success rate within a fixed budget of training steps. As illustrated in Fig. \ref{fig:sim_online}, our proposed method consistently demonstrates superior sample efficiency and asymptotic performance across both OGBench and RoboMimic tasks.

\begin{figure*}[ht]
    \centering
    \includegraphics[width=\textwidth]{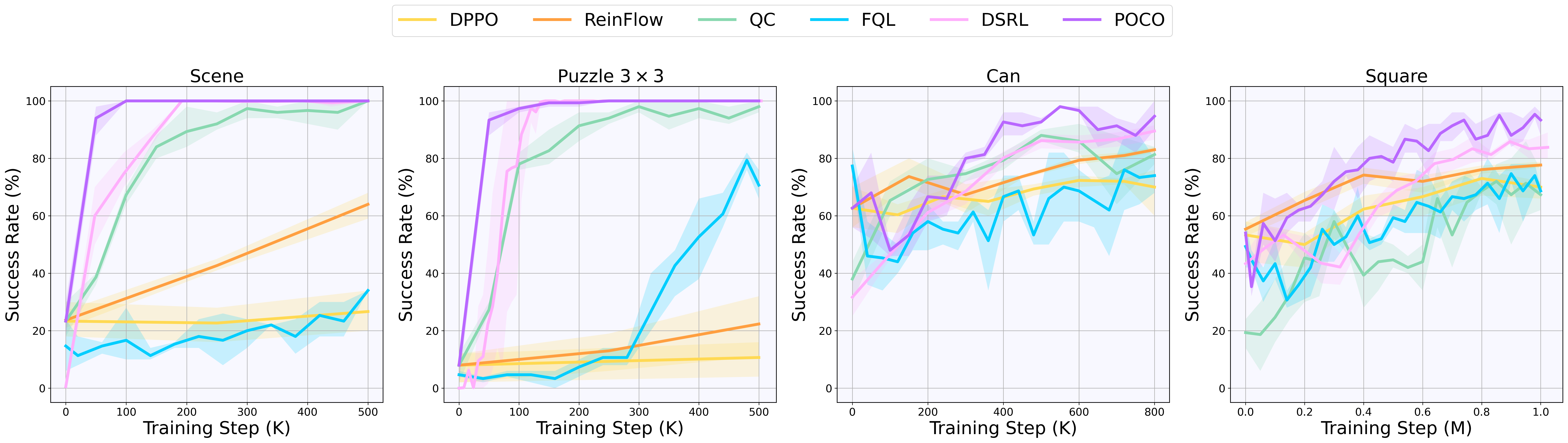}
    \caption{\textbf{Learning curves during offline-to-online training for 4 simulation tasks.} Results are averaged over 50 trials per task at each evaluation step with 5 random seeds.}
    \label{fig:sim_off2on}
\end{figure*}\textbf{}

\begin{table*}[t]
    \caption{\textbf{Policy training details for offline-to-online experiments.} }
    \label{tab:on2on_task_detail}
    \centering
    \resizebox{\textwidth}{!}{
    \begin{threeparttable}
    \begin{tabular}{c|cccc|cccc}
        \toprule
         &\multicolumn{4}{c}{\textbf{Simulation Tasks}} &\multicolumn{4}{c}{\textbf{Real-world Tasks}}\\
        Parameter &\texttt{Scene} &\texttt{Puzzle 3x3} &\texttt{Can} &\texttt{Square} &\texttt{Pick Cube} &\texttt{Route Cable} &\texttt{Insert USB} &\texttt{Assemble SSD}\\
        \hline 
        Batch size &256 &256 &256 &256 &256 &256 &256 &256\\
        Learning rate &3e-4 &3e-4 &3e-4 &3e-4 &3e-4 &3e-4 &3e-4 &3e-4\\
        $\gamma$ &0.99 &0.99 &0.99 &0.99 &0.99 &0.99 &0.99 &0.99\\
        $H$ &750 &500 &300 &400 &200 &150 &200 &180\\
        $T$ &5 &5 &5 &5 &4 &4 &4 &4\\
        Offline pre-trained step &500,000 &500,000 &100,000 &500,000 &20,000 &20,000 &20,000 &20,000\\
        Critic warmup step &5,000 &5,000 &5,000 &5,000 &5,000 &5,000 &5,000 &5,000\\
        $N$ &32 &32 &32 &32 &32 &32 &32 &32\\
        $\eta$ &0.001 &0.001 &0.001 &0.001 &0.001 &0.001 &0.001 &0.001\\
        $(\beta, \zeta)$ &$(1.0, 0.3)$ &$(1.0, 0.15)$ &$(1.0, 0.3)$ &$(1.0, 0.08)$ &$(1.0, 0.1)$ &$(1.0, 0.1)$ &$(1.0, 0.08)$ &$(1.0, 0.08)$\\
        EEF bounding box\tnote{1} &- &- &- &- &(40, 40, 30) &(20, 30, 30) &(20, 15, 15) &(12, 15, 15)\\
        Object random range\tnote{2} &- &- &- &- &(25, 10) &(5, 15) &(5, 5) &(5, 3)\\
        \bottomrule
    \end{tabular}
    \begin{tablenotes}
        \footnotesize
        \item[1] The unit of dimensions (x, y, z) is centimeter.
        \item[2] The unit of dimensions (x, y) is centimeter.
    \end{tablenotes}
    \end{threeparttable}
    }
\end{table*}

On simpler tasks such as \texttt{Scene}, \texttt{Puzzle 3x3}, and \texttt{Lift}, our method achieves high success rates significantly faster than the baselines. For instance, in the \texttt{Scene} task, POCO reaches a near 100\% success rate within 0.2M steps, whereas RLPD and QC require considerably more interaction to reach comparable performance, and FQL struggles to learn effectively under the sparse-reward case. This indicates that our framework efficiently leverages the offline dataset, and the chunked-Q accelerates backpropagation of sparse rewards, leading to rapid online convergence.

The performance gap widens on challenging tasks involving precise long-horizon manipulation, such as \texttt{Square}, \texttt{Cube-double}, and \texttt{Cube-triple}. In \texttt{Cube-triple}, which requires stacking 3 cubes, FQL completely fails to learn (near 0\% success), and the strong baseline RLPD achieves around 60\%. Since FQL relies on backpropagating gradients through the entire ODE solver process. In these long-horizon sparse-reward tasks, this leads to severe gradient degradation and the accumulation of value over-estimation, causing the policy to collapse early in training. In contrast, POCO steadily improves, achieving a near-100\% success rate. Similarly, in \texttt{Square}, POCO outperforms all baselines, demonstrating our method's ability to model complex multi-modal distributions that are difficult for direct gradient-based methods to capture.

While FQL and QC both utilize advanced generative models, they exhibit instability across different domains. FQL, which relies on direct Q-gradient backpropagation, exhibits high variance and fails on sparse-reward OGBench tasks, likely due to noise in Q-gradients during early training. QC performs reliably but learns slower than POCO, suggesting that POCO provides a more stable and efficient learning signal than Chunked-Q matching alone.

\subsection{Stability of Offline-to-Online Fine-tuning in Simulation}
\label{sec:off2on_sim}
To evaluate the learning stability of our approach, we compare QC\cite{qc}, FQL\cite{fql}, and 3 other competitive methods with our approach under an offline-to-online setting on 4 simulation tasks:
\begin{itemize}
    \item DSRL\cite{dsrl} that alters the input noise distribution instead of modifying the weights of the policy for performance improvement. 
    \item DPPO\cite{dppo}, which is an on-policy method that adapts the Proximal Policy Optimization framework to diffusion models.
    \item ReinFlow\cite{reinflow}, which is also an on-policy method that utilizes a noise injection network to improve the action generation of the  policy network.
\end{itemize}
The specific details of the policy training for each task are in Tab. \ref{tab:on2on_task_detail}. We report the average success rate within a fixed budget of training steps. Notably, for the on-policy methods (DPPO and ReinFlow), we deployed them using 20 parallel environments. To ensure a fair comparison, we align the metrics and normalize training steps by multiplying the number of update iterations by the number of parallel environments, maintaining a consistent total sample budget across all methods.

As illustrated in Fig. \ref{fig:sim_off2on}, POCO achieves the best balance of stability and sample efficiency, consistently avoiding catastrophic forgetting while reaching the highest asymptotic success rates across all 4 evaluated tasks. 

The on-policy methods, DPPO and ReinFlow, exhibit smooth and stable learning curves, confirming their ability to safely preserve the pre-trained prior. However, they suffer from sample inefficiency in these high-dimensional continuous control tasks. Particularly in environments like \texttt{Scene} and \texttt{Puzzle 3x3}, both methods struggle to make meaningful performance improvements within the allocated training step budget, rendering them impractical for real-world deployment.

\begin{figure*}[ht]
    \centering
    \includegraphics[width=\textwidth]{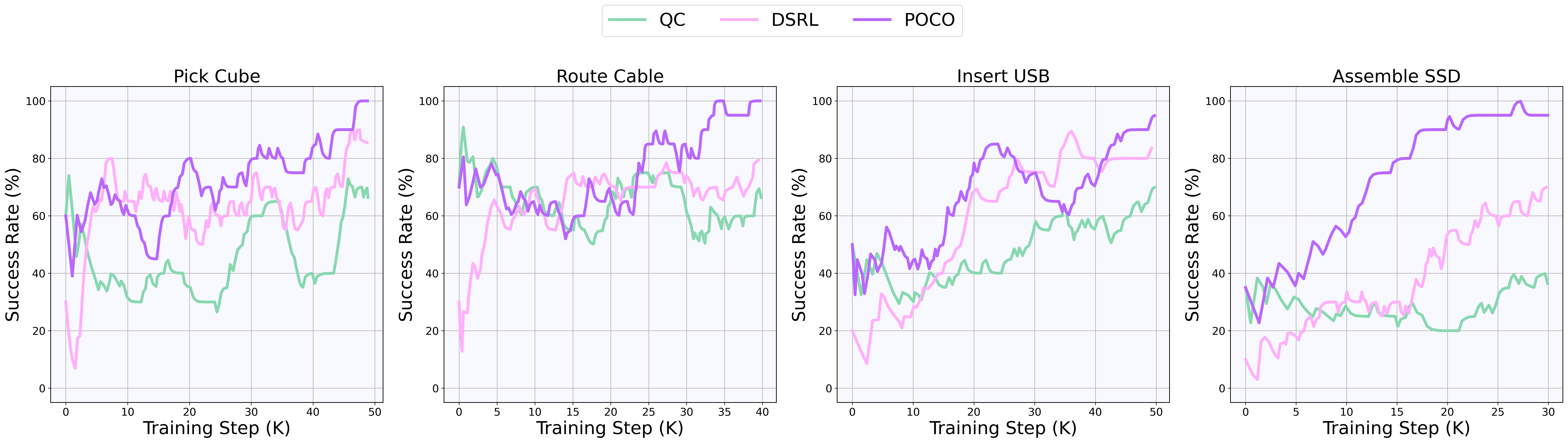}
    \caption{\textbf{Learning curves during offline-to-online training for 4 real-world tasks.} Results are displayed as a running average over 20 episodes.}
    \label{fig:real_online}
\end{figure*}

On the other hand, while off-policy methods like FQL and QC demonstrate higher sample efficiency, they exhibit distinct limitations. FQL is vulnerable to over-estimating OOD values, as evident in the \texttt{Square} task. While FQL completely fails to learn from scratch in the pure online setting, it manages to train in the offline-to-online setting only after an initial performance drop to near 35\%. This contrast demonstrates that early-stage inaccurate Q-values are fatal to direct gradient-based updates, as they instantly destroy the pre-trained behavioral prior before the critic can stabilize. Further, the supervised online update of QC is susceptible to sub-optimal exploratory rollouts. This causes instability and slow performance improvement during the online stage. 

DSRL avoids catastrophic forgetting by steering inference noise rather than updating policy weights, proving sample-efficient on simpler tasks like \texttt{Scene} and \texttt{Puzzle 3x3}. However, because its policy remains frozen at the pre-trained level, its capacity for continuous improvement is limited. Consequently, DSRL reaches its performance limit on complex, contact-rich tasks like \texttt{Can} and \texttt{Square}.

In contrast, POCO overcomes the limitations through the E-M formulation and robust clipped regression, which filters out noisy gradients from overestimated Q-values, preventing early catastrophic drops. Therefore, POCO safely updates policy weights, enabling continuous improvement.

\begin{table}[t]
\caption{\textbf{Success Rates of offline-to-online training for real-world tasks.} Results are reported over 30 trials per task.}
\label{tab:real_result}
\centering
\scalebox{1.0}{ 
    \begin{tabular}{c|cccc}
        \toprule
        \textbf{Task} &BC &QC &DSRL &POCO \\ 
        \midrule
        \texttt{Pick Cube}    &63.3\% &66.7\% &93.3\% &\textbf{100.0\%}  \\
        \texttt{Route Cable}  &73.3\% &70.0\% &80.0\% &\textbf{100.0\%}  \\
        \texttt{Insert USB}   &46.7\% &70.0\% &76.7\% &\textbf{90.0\%}   \\
        \texttt{Assemble SSD} &26.7\% &36.7\% &73.3\% &\textbf{96.7\%}   \\
        \rowcolor{gray!20} \textbf{Average}  &52.5\% &60.9\% &80.8\% &\textbf{96.7\%} \\
        \bottomrule
    \end{tabular}
}
\end{table}

\subsection{Real-World Robotic Manipulation}
\label{sec:real}

For the real-world benchmarks, we evaluate our algorithm on 4 sparse-reward robotic manipulation tasks of varying complexity. The specific details of the policy training for each task are in Tab. \ref{tab:on2on_task_detail}. To test the generalizability and robustness of the learned policies, these tasks are designed to challenge the agent across different physical dimensions: from basic spatial reasoning to complex contact-rich interactions. The selected tasks include: 1) \texttt{Pick Cube}, evaluating grasping and tight-tolerance alignment for a rigid cube; 2) \texttt{Route Cable}, assessing continuous force control over a deformable cable; 3) \texttt{Insert USB}, demanding asymmetric precision insertion; and 4) \texttt{Assemble SSD}, requiring a contact-rich oblique assembly process. We compare 2 competitive methods, QC and DSRL, with our approach under real-world settings. And we report the average success rate of a running average over 20 episodes within a fixed budget of training steps. As illustrated in Fig. \ref{fig:real_online} and quantified results in Tab. \ref{tab:real_result}, POCO consistently demonstrates the best online fine-tuning stability and final performance across all real-world tasks, achieving an average success rate of 96.7\%.

On tasks such as \texttt{Pick Cube} and \texttt{Route Cable}, our method, after offline pre-training, rapidly improves upon the pre-trained policy, achieving perfect execution. For instance, in the \texttt{Route Cable} task, POCO steadily increases its success rate to 100\% within 40K training steps. In contrast, QC struggles to achieve stable policy improvement within the limited training budget. Because QC directly uses a BC loss to update the policy, sub-optimal or OOD data generated during online rollouts may cause its performance to drop. Meanwhile, DSRL exhibits high sample efficiency and a stable learning performance, but its final performance remains bottlenecked by the initial offline-trained policy. This constraint restricts its capacity for further continuous improvement.

For tasks requiring high-precision, contact-rich execution, such as \texttt{Insert USB} and \texttt{Assemble SSD}, the advantages of POCO become even clearer. In the most challenging \texttt{Assemble SSD} task, which requires high-precision force-guided oblique insertion, the pre-trained policy only achieves a 30\% success rate. Due to sub-optimal rollout data, QC struggles to improve policy performance during online fine-tuning. Similarly, the sample efficiency of DSRL is also limited, resulting in a low success rate of 60\% within the same training step budget. POCO, however, exhibits a remarkably stable learning process and efficiently learns the optimal policy, achieving a 95\% success rate.

Overall, the real-world results highlight the practical limitations of current fine-tuning methods. Rather than explicitly filtering out over-estimated Q-values, POCO mitigates their destructive impact through two mechanisms: first, candidate actions are inherently grounded in the pre-trained prior distribution; second, the clipped surrogate objective strictly limits the gradient magnitude of any single action with an abnormal high weight. This creates a stable and highly efficient learning signal, allowing the agent to robustly improve its performance directly on a real-world platform.

\begin{figure}[t]
    \centering
    \includegraphics[width=\columnwidth]{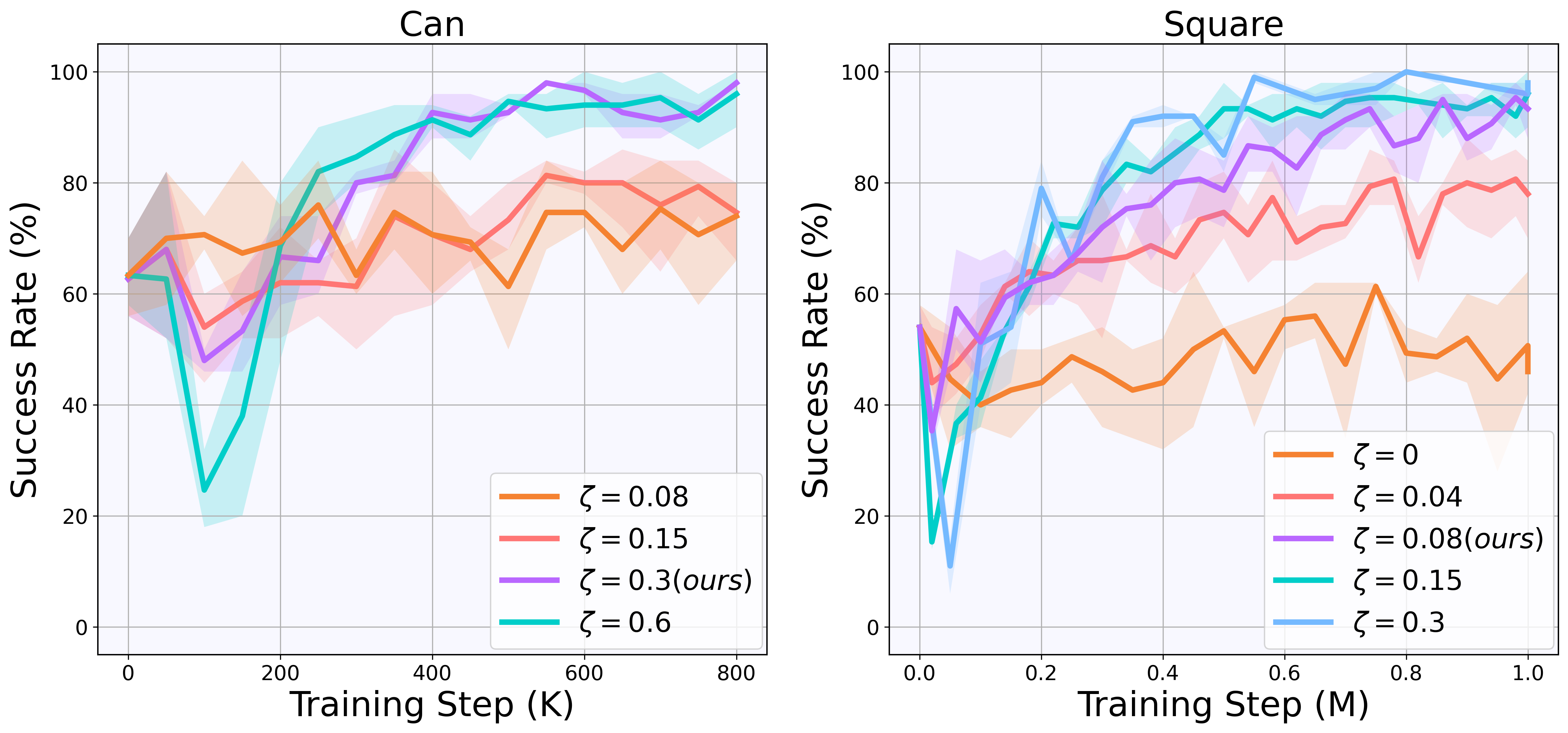}
    \caption{\textbf{Learning curves during offline-to-online training with different $\zeta$ for 2 simulation tasks.} Results are averaged over 50 trials per task at each evaluation step with 5 random seeds.}
    \label{fig:abla_clip}
\end{figure}

\subsection{Ablation Study}
\label{sec:ablation}
To evaluate the impact of key algorithmic designs in POCO, we conduct a series of ablation studies. Specifically, we investigate the sensitivity of our framework to two hyperparameters that govern the stability and efficiency of the policy update, the clipping threshold $\zeta$ and the posterior guidance scale $\beta$. By isolating these variables, we aim to understand how the trust-region constraint and the weighted posterior distillation influence the learning procedure. All ablation experiments are evaluated on representative simulation tasks to provide a clear and quantitative comparative analysis.
\subsubsection{Clipping Threshold $\zeta$}
The clipping threshold $\zeta$ bounds the maximum penalty applied to any candidate action within the surrogate objective, effectively controlling how strongly the posterior influences the policy update. As illustrated in Fig. \ref{fig:abla_clip}, selecting an appropriate $\zeta$ is crucial for balancing learning stability and improvement capacity. 

If $\zeta$ is set too large (e.g., $\zeta=0.3$ for \texttt{Square}), the clipped surrogate loss permits aggressively large parameter updates. During the early stages of the offline-to-online transition, the Q-function often suffers from inaccurate value estimates and over-estimation of OOD. Policy updates driven by these noisy Q-values rapidly destroy the pre-trained prior, causing the policy to collapse catastrophically, as evidenced by the sharp performance drop in the early stages. 

Conversely, if $\zeta$ is set too small (e.g., $\zeta=0$ or $\zeta=0.04$ for \texttt{Square}), the clip clipped surrogate strictly restricts the influence of the posterior, causing the update to be similar to standard supervised learning. As a result, it becomes difficult for the posterior to drive meaningful policy improvements, thereby limiting final performance. By establishing a balanced threshold (e.g., $\zeta=0.08$ for \texttt{Square}), POCO effectively enforces the trust-region constraint, preventing initial policy collapse while maintaining sufficient gradient flow to converge to the optimal performance.

\begin{figure}[t]
    \centering
    \includegraphics[width=\columnwidth]{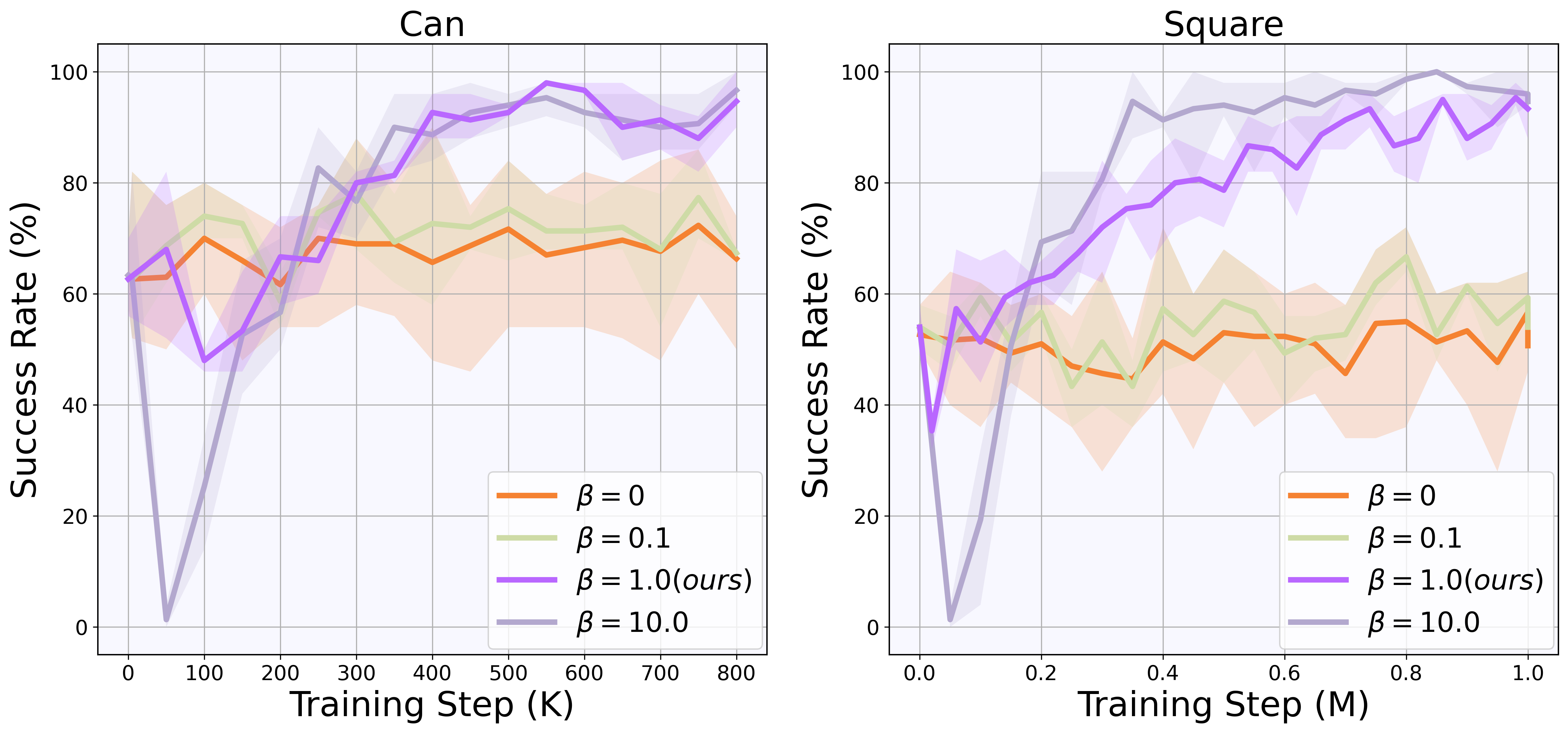}
    \caption{\textbf{Learning curves during offline-to-online training with different $\beta$ for 2 simulation tasks.} Results are averaged over 50 trials per task at each evaluation step with 5 random seeds.}
    \label{fig:abla_beta}
\end{figure}

\subsubsection{Posterior Guidance Scale $\beta$}
The posterior guidance scale $\beta$ controls the relative weight of the reward-weighted surrogate objective against the BC regularization. It essentially determines how strongly the inferred posterior influences the policy update. As illustrated in Fig. \ref{fig:abla_beta}, setting an appropriate $\beta$ is important for achieving effective and stable policy improvement.

When $\beta$ is set too small (e.g., $\beta=0.1$ for \texttt{Square}), the Q-value guidance is excessively weakened. Under this setting, the optimization objective largely degrades into standard supervised fine-tuning. Consequently, the agent fails to prioritize high-value actions effectively, resulting in minimal performance gains. Conversely, if $\beta$ is excessively large (e.g., $\beta=10.0$ for \texttt{Square}), the policy is forced to match the weighted posterior aggressively. During the early stages of online fine-tuning, the critic's Q-value estimates are often inaccurate. Over-reliance on this noisy guidance causes the policy to rapidly deviate from its pre-trained prior, leading to severe instability and catastrophic policy collapse, such as a sharp initial performance drop to near 0\% on the \texttt{Square} task. By adopting a balanced scale (e.g., $\beta=1.0$ for \texttt{Square}), POCO leverages the critic's guidance to continuously improve performance while maintaining sufficient regularization to ensure learning stability.

\begin{table}[t]
    \caption{\textbf{Policy training details for VLA scaling real-world experiments.} }
    \label{tab:task_detail_vla}
    \centering
    % \resizebox{\columnwidth}{!}{
    \begin{threeparttable}
    \begin{tabular}{c|cc}
        \toprule
        Parameter &\texttt{Pick Pen} &\texttt{Hang Keychain}\\
        \hline 
        Batch size &64 &64\\
        Learning rate for actor &5e-5 &5e-5\\
        Learning rate for critic &3e-4 &3e-4\\
        $\gamma$ &0.99 &0.99\\
        $H$ &200 &200\\
        $T$ &10 &10\\
        Critic warmup step &2,000 &2,000\\
        $N$ &16 &16\\
        $\eta$ &0.001 &0.001\\
        $(\beta, \zeta)$ &$(1.0, 0.01)$ &$(1.0, 0.01)$\\
        EEF bounding box\tnote{1} &(40, 40, 40) &(40, 40, 40)\\
        Object random range\tnote{2} &(30, 30) &(20, 20)\\
        \bottomrule
    \end{tabular}
    \begin{tablenotes}
        \footnotesize
        \item[1] The unit of dimensions (x, y, z) is centimeter.
        \item[2] The unit of dimensions (x, y) is centimeter.
    \end{tablenotes}
    \end{threeparttable}
    % }
\end{table}

\begin{table}[t]
\caption{\textbf{Success Rates of offline-to-online training for 2 real-world tasks with VLA models.} Results are reported over 30 trials per task. }
\label{tab:real_result_vla}
\centering
% \resizebox{\columnwidth}{!}{
\begin{threeparttable}
    \begin{tabular}{c|cc|cc}
        \toprule
        \textbf{Base Model} &\multicolumn{2}{c}{$\pi_{0.5}$} &\multicolumn{2}{c}{GR00T N1.6}\\
        \textbf{Task} &SFT &POCO\tnote{1} &SFT &POCO\tnote{1} \\ 
        \midrule
        \texttt{Pick Pen}      &76.7\% &\textbf{93.3\%} &63.3\% &86.7\% \\
        \texttt{Hang Keychain} &60.0\% &\textbf{86.7\%} &53.3\% &83.3\% \\
        \bottomrule
    \end{tabular}
    \begin{tablenotes}
        \footnotesize
        \item[1] Results for POCO are reported within 15,000 training steps.
    \end{tablenotes}
\end{threeparttable}
% }
\end{table}

\subsection{Scalability to VLA Models}
\label{sec:vla}

For the VLA scalability experiments, we adapt the aforementioned model setup by substituting the actor with the pre-trained VLA models. Meanwhile, the architecture of the critic network remains unchanged, retaining the same policy network for stable value estimation during the online fine-tuning procedure. The specific details of the policy training for each task are in Tab. \ref{tab:task_detail_vla}.

To evaluate the scalability and model-agnostic nature of our approach, we select two representative generative VLAs, $\pi_{0.5}$ \cite{pi05} and GR00T N1.6 \cite{gr00tn1}, as our base models. We select two real-world tasks as the benchmark: 1) \texttt{Pick Pen}, evaluating precise grasping and spatial alignment for a thin object; and 2) \texttt{Hang Keychain}, assessing manipulating a partially deformable object to a target hook. 

To ensure training efficiency and balance computational resources while fully leveraging the extensive pre-trained knowledge of these large-scale VLAs, we adopt a parameter-efficient fine-tuning strategy. Specifically, we freeze the encoders and the transformer backbones, restricting the parameter updates to the flow-based action head. Furthermore, to optimize the computational bottleneck during the Implicit E-step, which requires sampling multiple candidate actions, we execute the transformer forward pass only once per state. The resulting conditional state embeddings, or KV cache, are then reused by the action head to generate $N$ candidate action chunks. This architectural design drastically reduces the inference overhead, enabling high-frequency policy updates during online rollouts.

As detailed in Tab. \ref{tab:real_result_vla}, purely offline fine-tuning struggles to achieve high performance in precision-demanding tasks, even with the powerful visual and semantic priors embedded in large-scale VLA models. In the \texttt{Pick Pen} task, the offline fine-tuned $\pi_{0.5}$ model achieves a 76.7\% success rate, whereas GR00T N1.6 only reaches 63.3\%. This highlights the inherent limitation of supervised learning: without feedback from interactive environments, large models still struggle to overcome sub-optimal states and spatial misalignments during physical interaction.

By continuously improving the policy through the E-M procedure, POCO boosts the performance of these VLAs on downstream tasks. For example, POCO pushes the success rate of $\pi_{0.5}$ on \texttt{Pick Pen} from 76.7\% to 93.3\%. To sum up, these results demonstrate that POCO scales seamlessly to large network architectures. It serves as an efficient and stable alignment mechanism, unlocking the full potential of pre-trained foundational models for robust deployment in complex real-world environments.

\section{Conclusion}
In this paper, we introduce Posterior Optimization with Clipped Objective (POCO), a principled RL framework that formulates policy improvement as a likelihood-free posterior inference problem. By employing a robust clipped regression objective, POCO combines off-policy efficiency with strong prior regularization, effectively mitigating catastrophic policy collapse during generative policy fine-tuning. Evaluations across simulation and real-world benchmarks demonstrate that POCO achieves SOTA sample efficiency and stable improvement, while seamlessly scaling to fine-tune large VLA models for robust real-world deployment.

While POCO offers a reliable training mechanism, exciting avenues for future work remain. First, transitioning from random exploration to prior-driven, structured exploration may further improve sample efficiency in complex, contact-rich tasks. Second, since POCO relies on accurate value estimation, incorporating automated dense reward generation, such as world-model-driven reward shaping, is a critical next step to facilitate robust critic learning. Combining meaningful exploration and dense rewards is essential to advance our framework toward fully autonomous, open-ended robot learning.

\bibliographystyle{IEEEtran}
% \bibliography{references}

\appendices

\section{Mathematical Proofs}
\subsection{The Closed Form of the Optimal Variational Posterior}
\label{apx:closed_form_q}
We hereby derive the closed-form solution for the optimal variational distribution $q(a|s)$ presented in Eq. \ref{eq:closed_form_q}. 

In the E-step, the policy parameters are fixed at $\theta = \theta_i$. We seek the distribution $q(a|s)$ that maximizes the objective. By definition, the expected cumulative discounted return starting from a state $s$ is captured by the state-action value function $Q_{\pi_i}(s, a)$. Therefore, maximizing the global cumulative objective in Eq. \ref{eq:elbo_final} is equivalent to maximizing the local objective for each visited state $s$ independently:
\begin{equation}
    \mathcal{J}_{\text{E-step}}(q)=\mathbb{E}_{a\sim q(\cdot|s)}[Q_{\pi_i}(s, a)]-\eta D_{\text{KL}}(q(\cdot|s)||\pi(\cdot|s,\theta_i))
\end{equation}

We define the Lagrangian $\mathcal{L}(q, \lambda)$ to enforce the normalization constraint $\int q(a|s) da = 1$:
\begin{equation}
    \begin{aligned}
    \mathcal{L}(q,\lambda)=&\int q(a|s)Q_{\pi_i}(s,a)da \\
    &-\eta\int q(a|s)\Big(\log\frac{q(a|s)}{\pi(a|s,\theta_i)}\Big)da \\
    &+\lambda\Big( \int q(a|s)da-1\Big)
    \end{aligned}
\end{equation}

Taking the functional derivative with respect to $q$:
\begin{equation}
    \begin{aligned}
    \frac{\delta \mathcal{L}}{\delta q}&=Q_{\pi_i}(s, a)-\eta(\log q(a|s)+1) +\eta\log \pi(a|s,\theta_i)+\lambda 
    \end{aligned}
\end{equation}
Setting it to zero and then we have:
\begin{equation} 
    \eta\log q(a|s)=Q_{\pi_i}(s, a)+\eta\log\pi(a|s,\theta_i)+(\lambda-\eta) 
\end{equation}

Dividing by $\eta$ and exponentiating both sides:
\begin{equation}
    q(a|s)=\pi(a|s,\theta_i)\exp\Big(\frac{Q_{\pi_i}(s,a)}{\eta}\Big)\exp\Big(\frac{\lambda-\eta}{\eta}\Big)
\end{equation}

The term $\exp((\lambda-\eta)/\eta)$ is constant with respect to action $a$ and serves as the normalization constant. Thus, we have the closed-form solution:
\begin{equation}
    q(a|s) \propto \pi(a|s, \theta_i) \exp\Big( \frac{Q_{\pi_i}(s, a)}{\eta} \Big)
\end{equation}

\vfill

\end{document}